\documentclass[final]{opt2020}

\usepackage[utf8]{inputenc} 
\usepackage[T1]{fontenc}    
\usepackage{hyperref}       
\usepackage{url}            
\usepackage{booktabs}       
\usepackage{amsfonts}       
\usepackage{nicefrac}       
\usepackage{microtype}      
\usepackage{faktor}
\usepackage{hhline}

\usepackage{amsthm}
\usepackage{amsmath}
\usepackage{amssymb}
\usepackage{wrapfig}

\usepackage{tikz}

\title{A Study of Condition Numbers for First-Order Optimization}

\optauthor{\Name{Charles Guille-Escuret}\thanks{equal contributions} \Email{charles.guille-escuret@umontreal.ca}\\
\addr Mila, Universit\'e de Montr\'eal
\AND
\Name{Baptiste Goujaud}\footnotemark[1] \Email{baptiste.goujaud@gmail.com}\\
\addr Mila
\AND
\Name{Manuela Girotti} \Email{manuela.girotti@umontreal.ca}\\
\addr Mila, Universit\'e de Montr\'eal, Concordia University
\AND
\Name{Ioannis Mitliagkas} \Email{ioannis@iro.umontreal.ca}\\
\addr Mila, Universit\'e de Montr\'eal, Canada CIFAR AI chair}

\theoremstyle{plain}
\newtheorem{Th}{Theorem}[section]

\newtheorem{Cor}[Th]{Corollary}
\newtheorem{Prop}[Th]{Proposition}

\theoremstyle{definition}
\newtheorem{Def}[Th]{Definition}

\newtheorem{Rem}[Th]{Remark}
\newtheorem{?}[Th]{Problem}
\newtheorem{Ex}[Th]{Example}

\def\SC{\operatorname{SC}}
\def\EB{\operatorname{EB}}
\def\RSI{\operatorname{RSI}}
\def\PL{\operatorname{PL}}
\def\QG{\operatorname{QG}}

\def\R{\mathbb{R}}
\def\d{{\rm d }}

\def\le{\left}
\def\ri{\right}

\begin{document}
    \maketitle
    
    \begin{abstract}
        
The study of first-order optimization algorithms (FOA) typically starts with assumptions on the objective functions, most commonly smoothness and strong convexity. These metrics are used to tune the hyperparameters of FOA. We introduce a class of perturbations quantified via a new norm, called *-norm. We show that adding a small perturbation to the objective function has an equivalently small impact on the behavior of any FOA, which suggests that it should have a minor impact on the tuning of the algorithm. However, we show that smoothness and strong convexity can be heavily impacted by arbitrarily small perturbations, leading to excessively conservative tunings and convergence issues. In view of these observations, we propose a notion of continuity of the metrics, which is essential for a robust tuning strategy.
Since smoothness and strong convexity are not continuous, we propose a comprehensive study of existing alternative metrics which we prove to be continuous. We describe their mutual relations and provide their guaranteed convergence rates for the Gradient Descent algorithm accordingly tuned. Finally we discuss how our work impacts the theoretical understanding of FOA and their performances.


    \end{abstract}
    
    \section{Introduction}\label{sec:introduction} 
    Optimization of a high-dimensional cost function is at the core of fitting most machine learning models. In practice this is almost always performed by gradient-based first-order optimization algorithms (FOA). The analysis of their convergence properties typically assumes that their hyper-parameters are tuned based on some function properties; for example it is well-known that if $f$ is $\mu$-strongly convex and $L$-smooth, then gradient descent (GD) with step size $\alpha=\frac{2}{\mu+L}$ achieves a global linear convergence rate of $1-\frac{2}{\kappa+1}$ where $\kappa=\frac{L}{\mu}$ is called the condition number (see e.g. \citep{Nesterov2}). The condition number gives an indication of the tightness of the bounds on the curvature of $f$, and therefore of the difficulty to optimize it: the bigger the value of $\kappa$ is, the slowest is the convergence of the algorithm.

In \citep{Lessard}, the authors introduce a piece-wise quadratic function $f_{\rm LRP} \in C^1(\R)$, with a smaller second-order derivative for $x\in [1, 2]$ than elsewhere. They show that the Heavy-Ball (HB) algorithm \citep{Polyak}, when tuned using the $L$ of smoothness and the $\mu$ of strong convexity of $f_{\rm LRP}$, does not converge to the unique absolute minimizer $x=0$ for some initialization $x_0>0$. On the other hand, standard GD with step size $\alpha=\frac{2}{\mu+L}$ does converge, but at a very low rate due to the high condition number of $f_{\rm LRP}$. 
Although tuning the HB algorithm based on the  $L$-smoothness and $\mu$-strong convexity of $f_{\rm LRP}$ is arguably a heuristic strategy (since this optimal tuning rule is only provided for \textit{quadratic} functions, 
see \cite{Polyak_opt}),  the example in \citep{Lessard} highlights a striking pathological behaviour: a localized, bounded perturbation of the Hessian of the objective function yields a disastrous effect on its condition number and on the trajectory of the iterates of the FOA.

In this paper we analyze this phenomenon and we propose a unifying framework to study the convergence of FOA and design robust tunings of their hyperparameters. We first introduce a new topology, based on the definition of a star-norm $\le\|\cdot\ri\|_*$  (Section \ref{subsec:starnorm}). Such a norm will be the fundamental tool that will be used throughout the paper in order to assess the ``closeness'' between objective functions: Theorem \ref{continuity1} states that two functions whose difference is small in the $\|\cdot\|_*$-norm sense have comparable behaviour under continuous FOA. 
Therefore, the tuning strategy for the hyperparameters of the optimization algorithm should account for this similarity. However, the standard tuning based on smoothness and strong convexity fails to do so (Theorem \ref{continuity4}) and it is easy to construct examples that illustrate this weakness.

Based on such a topology, we then define the notion of continuity of a condition number (Section \ref{cond_number_cont}), which in turn reflects the continuity of some properties of the objective function that we call \textit{upper/lower conditions} (Section \ref{sec:notations}), smoothness and strong convexity being two examples of them. Having a continuous condition number is essential to the robustness of both tuning methods and convergence rates of the FOA.
Our approach implies that even when the objective function verifies some of the strongest conditions (strong convexity and smoothness), relying on weaker ones to tune the FOA can lead to better and more consistent convergence behaviours.

\section{Related Work}\label{work}

Because strong convexity and smoothness are strong requirements that are not verified by some classic machine learning models such as logistic regression (which verifies convexity but not strong convexity), several works have already explored substitute assumptions. Alternatives to strong convexity, which we will call \emph{lower conditions} have been the most thoroughly studied, under some overlapping names. These include \emph{local-quasi-convexity} \citep{NIPS2015_5718}, \emph{weak quasi-convexity} \citep{hardt2016gradient}, \emph{restricted secant inequality} ($\RSI$) \citep{zhang2013gradient}, \emph{error bounds} ($\EB$) \citep{luo_error_1993}, \emph{quadratic growth} ($\QG$) \citep{10.1137/S1052623499359178}, \emph{Polyak-{\L}ojasiewicz} ($\PL$) \citep{polyak_gradient_1963}, further generalized as \emph{Kurdyka-{\L}ojasiewicz} ($\operatorname{KL}$) \citep{kurdyka1998gradients}\citep{bolte2008characterizations}. The scattering of these notions in the literature has led to some confusing names. For example, \emph{optimal strong convexity} ($\operatorname{OSC}$) \citep{liu2014asynchronous} is also called \emph{semi-strong convexity} \citep{gong2014linear} and \emph{weak strong convexity} \citep{ma2015linear}, despite being a different notion from the \emph{weak strong convexity} of \citep{DBLP:journals/corr/KarimiNS16}, which was formerly called \emph{quasi-strong convexity} ($\operatorname{QSC}$) in \citep{necoara2015linear}. Similarly, the \emph{restricted strong-convexity} from \citep{agarwal2011fast} is a different notion from the \emph{restricted strong convexity} of \citep{zhang2013gradient}. To avoid further confusion, we will use the name \emph{star-strong convexity} ($^*\!\SC$) for the notion of $\operatorname{WSC}$/$\operatorname{QSC}$ of \citep{DBLP:journals/corr/KarimiNS16}.

Alternatives to smoothness, which we will call \emph{upper conditions}, have also been proposed, though more sporadically, such as \emph{local smoothness} \citep{NIPS2015_5718}, \emph{restricted smoothness} \citep{agarwal2011fast}, \emph{relative smoothness} \citep{lu2016relativelysmooth}, \citep{hanzely2018accelerated}, \citep{zhou2019simple}, \emph{restricted Lipschitz-continuous gradient} ($\operatorname{RLG}$) \citep{zhang2013gradient}.

Most \emph{lower conditions} can naturally be translated into an equivalent \emph{upper condition}, by shifting the inequality from a lower bound to an upper bound. For example, smoothness is an upper condition equivalent to strong convexity, and \emph{weak-smoothness} \citep{hardt2016gradient} is an upper condition equivalent of the PL condition, which is further generalized to the stochastic case as \emph{expected smoothness} in \citep{gower2019sgd}. Similarly, $\RSI$, $\operatorname{WSC}$, $\EB$, and $\QG$ all have natural equivalent \emph{upper conditions}. In an attempt to reduce the number of similar names and their associated confusion, we will for instance refer to the PL condition as $\PL^-(\mu)$ and to its equivalent \emph{upper condition} as $\PL^+(L)$. 

In \citep{DBLP:journals/corr/KarimiNS16} the authors propose a study of the implications between some \emph{lower conditions}, although under the assumption of global smoothness, and omitting the constant conversion induced by the implications. To the best of our knowledge, a study of the implications between \emph{upper conditions} is missing from the literature.  
 We collect all the relations between \emph{upper} and \emph{lower} conditions in two implication graphs (\figurename\ \ref{assumptions_graph}), together with the constant conversions. We also study upper bounds on the convergence rates of gradient descent assuming  that the objective function satisfies each pair of upper/lower conditions (Table \ref{convergence_table}).

 While alternative conditions have been extensively researched, the main goal  of the works mentioned above has always been to extend convergence results to a larger class of functions. On the other hand, our work aims at introducing a new approach for tackling the optimization task and at bringing a deeper understanding on the convergence of FOA and its connection with properties of the objective function itself.


\section{Setup and notation}\label{notation}




In this paper, we focus on minimizing an objective function $f:\R^d\to\R$ using first-order algorithms (FOA).
The objective function is assumed to be continously differentiable $f\in C^1(\R^d)$, with a convex set of global minima $X^*\subseteq\mathbb{R}^d$;
we denote $f^*= \min_{x\in\mathbb{R}^d}f(x)$.
For $x\in\mathbb{R}^d$, we denote the distance between $x$ and $X^*$ as $d(x,X^*)=\inf_{x^*\in X^*}\|x-x^*\|_2$. We recall that since $X^*$ is convex, for every $x\in \R^d$ there exists a unique element $x^*_p \in X^*$ (called the projection of $x$ onto $X^*$) such that $ \|x-x^*_p\|_2 = d(x,X^*)$.

For our analysis we will consider the following class of deterministic FOA:
\begin{Def}[\bf Continuous FOA]
A first-order algorithm $\mathcal{A}_\theta$, possibly depending on a set of hyperparameters $\theta$, is continuous if  $\forall \, n \in \mathbb{N}$ the $(n+1)$-iterate 
    \begin{gather}
        x_{n+1}=
    \mathcal{A}_\theta\Big( \{x_{i}\}_{i=0\ldots n}, \{f(x_i)\}_{i=0\ldots n}, \{\nabla f(x_i)\}_{i=0\ldots n} \Big), 
    \end{gather}
    is continuous with respect to all of its arguments.

\end{Def}
    Trivially, any algorithm that can be expressed as a finite composition of continuous operations is continuous. This class of FOA includes all the major algorithms like GD and Heavy Ball (HB) methods with step size and momentum hyperparameters not depending on the local values of $f$.
    However, some methods like Polyak step size \citep{Polyak_opt} are not guaranteed to be continuous without additional assumptions on the objective function.

We denote $\mathcal{B}(X^*,r) = \{y\in\mathbb{R}^d |\: d(y,X^*) < r\}$ to be the set of points in $\R^d$ whose distance from set $X^*$ is smaller than  $r$ and, for a set of functions $\mathcal{F}$, we denote $f+\mathcal{F}$ the set of functions $g$ such that $g-f \in\mathcal{F}$.

Unless stated otherwise, rates of convergence refer to the convergence of $f(x_n)-f^*$, not $d(x_n,X^*)$.

    \section{Continuity of first-order algorithms and condition numbers}\label{sec:continuity}

In this section  we introduce the theoretical framework to analyze the behaviour of FOA for objective functions that are "close". 
The first necessary component is a norm $\|\cdot\|_*$ that will induce the right kind of topology to evaluate the similarity between objective functions. 
Proofs of all key results are collected in Appendix \ref{App_continuity}.

\subsection{Star norm and stability of FOA behaviors}\label{subsec:starnorm}

Consider an objective function $f\in C^1(\R^d)$. The purpose of the $\|\cdot\|_*$-norm  will be to evaluate the impact of a perturbation of $f$ on the convergence properties of FOA. In particular, if two functions $f$ and $g$ are such that $\|f-g\|_*$ is small, it is desirable for the FOA to behave similarly on them. 
Since we are focussing on optimization algorithms that depend on the first derivatives of the function, we require the $\|\cdot\|_*$-norm to give some control over the amplitude of the gradient of the perturbation of $f$. Additionally, notice that as the iterates approach the minima of the objective function, the updates typically become finer, so that even a small perturbation of the function gradient can greatly affect the convergence behaviour. This supports the intuition that the same perturbation of the gradient will have more impact close to the set of minima $X^*$, and less impact far away. 

In view of the above discussion, we introduce the following definition of the $\|\cdot\|_*$-norm, which measures the maximal perturbation of the gradient weighted by the inverse of the distance to $X^*$.

\begin{Def}[\bf Star norm]\label{topology_*norm}
    Let $X^*\subseteq\mathbb{R}^d$ and 
    \begin{align*}
        \mathcal{F}_{X^*}=\{h\in C^1(\mathbb{R}^d)\mid \forall \, x^*\in X^*, h(x^*)=0
        \text{ and }  \exists \, L\in\mathbb{R}: \|\nabla h(x)\|_2\leq L\ d(x,X^*), \forall \, x\in\mathbb{R}^d\}.
    \end{align*}
    We define the {\em star norm}, $\le\|\cdot\ri\|_*$, on $\mathcal{F}_{X^*}$ as 
    \begin{gather}
    \forall \, h\in\mathcal{F}_{X^*}, \quad \|h\|_*= \sup_{x\in\mathbb{R}^d\backslash X^*}\frac{\|\nabla h(x)\|_2}{d(x,X^*)}.
    \end{gather}
\end{Def}

\begin{Ex}
Consider the function $h(x) = \sqrt{x^2+1}-1$, which can be thought as a differentiable version of the absolute value function; then, $h\in \mathcal{F}_{\{0\}}$, with $\|h\|_*=1$.  
\end{Ex}

\begin{Ex}
Similarly, consider the function 
\begin{gather}
h(x) = \begin{cases}
0 & x<1\\
(x-1)^3& 1\leq x\leq 2\\
3x-5 & x>2
\end{cases}
\end{gather}
then  $h\in \mathcal{F}_{\{0\}}$, with $\|h\|_*=\frac{3}{2}$.
\end{Ex}

\begin{Rem}
We emphasize that neither $X^*$ nor $\mathcal{F}_{X^*}$ depend of the objective function $f$, which does not need to be in $\mathcal{F}_{X^*}$ itself. Requiring $h(x^*)=0$ ensures that the $\|\cdot\|_*$-norm is indeed a norm on $\mathcal{F}_{X^*}$. Equivalently, we could have considered the quotient space $\faktor{\mathcal{F}_{X^*}}{[c]}$, where $[c]$ is the set of constant functions, equipped with $\|\cdot\|_*$; however, this would have introduced too many technicalities along the paper, therefore we did not proceed in this direction.
\end{Rem}


Let $x_i(f,\mathcal{A}_\theta,x_0)$ denote the $i$-th iterate obtained by applying a prescribed algorithm $\mathcal{A}_\theta$ to $f$ starting in $x_0$.
We now argue that two functions that are close in the sense of the star norm will have similar behaviors for continuous FOA.

\begin{Th} \label{continuity1}
    Let $f\in C^1(\R^d)$ with a set of global minimizers $X^*$ and $\le\|\cdot\ri\|_*$ the corresponding star norm. Let $\mathcal{A}_\theta$ be a continuous first-order algorithm and $\mathcal{K}\subset \mathbb{R}^d$ a compact set. Then, the following result holds:
    \begin{gather*}
        \forall \, \epsilon >0, \ \forall \, i\in \mathbb{N}, \exists \, \eta= \eta(\epsilon, i, \mathcal{K})>0  \text{ such that } \\
        \forall \, h \in \mathcal{F}_{X^*}, \text{ if } \|h\|_*<\eta, \text{ then }\\ \forall \, x_0\in\mathcal{K}: \|x_i(f, \mathcal{A}_\theta, x_0) - x_i(f + h,\mathcal{A}_\theta, x_0) \|_2<\epsilon.
    \end{gather*}
\end{Th}



The following corollary proves that for a target neighborhood of $X^*$ and any $\delta>0$, if $h$ is sufficiently small in the sense of $\le\|\cdot\ri\|_*$, then $\forall \, x_0\in\mathcal{K}$, applying $\mathcal{A}_\theta$ to $f + h$ starting in $x_0$ will attain the target neighborhood in exactly the same number of steps as for $f$, up to a distance tolerance of $\delta$.

\begin{Cor} \label{continuity3}
    Under the same hypotheses as Theorem \ref{continuity1}, let $\varepsilon>0$ and $\mathcal{B}(X^*,\varepsilon)$ a target neighborhood of $X^*$. Let us assume that $\mathcal{A}_\theta$ applied to $f$ converges to $X^*$ and $\forall \, x_0 \in \mathcal{K}$, let $N_{x_0} \in \mathbb{N}$ the smallest number of iterations such that 
    $x_{N_{x_0}} (f,\mathcal{A}_\theta,x_0) \in \mathcal{B}(X^*,\varepsilon).$ Then,\\ $\forall\, \delta>0$, $\exists\, \eta>0$ s.t. for any $h\in\mathcal{F}_{X^*}$, if $\|h\|_*< \eta$, then $\forall \, x_0\in\mathcal{K}$, 
    \begin{gather*}
    x_{N_{x_0}-1}(f+h,\mathcal{A}_\theta,x_0)\notin \mathcal{B}(X^*, \varepsilon-\delta) \quad \text{and} \quad
        x_{N_{x_0}}(f+h,\mathcal{A}_\theta,x_0)\in \mathcal{B}(X^*, \varepsilon+\delta).
    \end{gather*}
\end{Cor}

Theorem \ref{continuity1} and Corollary \ref{continuity3} show that if $h$ is sufficiently small in the sense of the norm $\|.\|_*$, then the behaviour of a continuous FOA on $f$ and $f + h$ will be similar, and thus it is natural to assume that the tuning of hyperparameters $\theta$ should also be similar. 
However, as the next section shows, this is not always the case.

\subsection{Standard tuning fails continuity test}\label{fail_tuning}
Consider the family of piecewise quadratic functions $\{f_\varepsilon\}_{\varepsilon\geq 0} \subset C^1(\R)$: 
\begin{gather}
f_\varepsilon (x) = \begin{cases}
x^2 &   x\leq 1\\
x^2+\frac{1}{\varepsilon}x^2-\frac{2x}{\varepsilon}+\frac{1}{\varepsilon} & 1\leq x\leq 1+\varepsilon^2\\
x^2+2\varepsilon x-2\varepsilon-\varepsilon^3 & x\geq 1+\varepsilon^2
\end{cases}
\end{gather}
We can view each function $f_\varepsilon$ as a perturbation of the quadratic $f_0(x)= x^2$, which is $2$-smooth and $2$-strongly convex: $\forall \, \varepsilon \geq 0$, $f_\varepsilon(x) = f_0(x) + h_\varepsilon(x)$ with $h_\varepsilon \in \mathcal{F}_{X^*=\{0\}}$. It is also  easy to see that $\| h_\varepsilon \|_* \to 0$ as $\varepsilon \to  0$.

The following properties hold:
\begin{Prop}
\label{prop-continuity}
    For any $\varepsilon> 0$, the function $f_\varepsilon$ is $\mu_\varepsilon$-strong  convex and $L_\varepsilon$-smooth, with $\mu_\varepsilon = 2$ and $L_\varepsilon = 2+\frac{2}{\varepsilon}$; moreover,  these constants are optimal: i.e. $f_\varepsilon$ is not $\mu$-strongly convex for $\mu>2$ and not $L$-smooth for $L<2+\frac{2}{\varepsilon}$. 
    
    Furthermore, GD tuned with step size $\alpha= \frac{2}{\mu_0+ L_0} = \frac{1}{2}$ applied to $f_\varepsilon$ ($\forall \, \epsilon > 0$) converges with linear rate $\varepsilon$; however, if GD is tuned with $\alpha=\frac{2}{\mu_\varepsilon+L_\varepsilon}=\frac{\varepsilon}{2\varepsilon+1}$, it does not converge with linear rate $q$ for any $q< \frac{1-\varepsilon^2}{(2\varepsilon+1)(1+\varepsilon^2)}$.
\end{Prop}

If we tune GD according to the values of smoothness and strong convexity of $f_0$
and optimize $f_\varepsilon$, the linear rate tends to $0$ as $\varepsilon \to 0$ (in fact, we obtain convergence in at most two steps). On the other hand, if we tune GD based on the tightest strong convexity and smoothness constants  $\mu_\varepsilon$ and $L_\varepsilon$ of $f_\varepsilon$, the linear rate tends to $1$ as $\varepsilon \to 0$. 
Notice that the condition number $\frac{L_\varepsilon}{\mu_\varepsilon}$ of $f_\varepsilon$ diverges as $\varepsilon \to 0$, thus leading to a very conservative tuning and increasingly slow convergence rate, while the tuning of $f_0$ leads to superlinear convergence.

The above example suggests that a sane tuning strategy for the hyperparameters of a FOA should be robust (continuous) with respect to $\|\cdot\|_*$-small perturbations of a given function.
It also shows that the standard tuning based on $L$-smoothness and $\mu$-strong convexity lacks this property.

%
%

\subsection{Continuity of condition numbers}\label{cond_number_cont}

We now formally introduce the notions of upper and lower conditions which represent generalizations of smoothness and strong convexity, and the notion of continuity of a condition.

\begin{Def}[\bf Upper conditions]
We use the term {\em upper condition} to describe a generalization of smoothness and we formalize it as a family of sets of functions, $\mathcal{C}^+(L)\subseteq C^1(\mathbb{R}^d)$, which satisfies 
$\mathcal{C}^+(L_1)\subseteq\mathcal{C}^+(L_2)$ for all $ L_1\leq L_2$.
\end{Def}
\begin{Def}[\bf Lower conditions]
We use the term {\em lower condition} to describe a generalization of strong convexity. 
We formalize it as a family of sets of functions, $\mathcal{C}^-(\mu)\subseteq C^1(\mathbb{R}^d)$, which satisfies 
$\mathcal{C}^-(\mu_1)\supseteq\mathcal{C}^-(\mu_2)$ for all  $\mu_1\leq\mu_2$.
\end{Def}
In Definition~\ref{lower-bounds-properties} and Definition~\ref{upper-bounds-properties}, we list some known upper and lower conditions extensively studied in the literature .
    
\begin{Def}[\bf Continuity of a condition] \label{condition_continuity}
    We say that $\mathcal{C}^+$ is continuous in $f\in\bigcup_{L>0}\mathcal{C}^+(L)$ with convex set of global minima $X^*$ if for any $L>0$ s.t. $f\in \mathcal{C}^+(L)$, $\forall\, \varepsilon>0, \exists \, \eta>0$ s.t. $\forall \, h\in \mathcal{F}_{X^*}, \text{if } \|h\|_* \leq \eta$, then $f+h\in\mathcal{C}^+(L+\varepsilon)$. 
    
    Similarly, $\mathcal{C}^-$ is continuous in $f\in\bigcup_{\mu>0} \mathcal{C}^-(\mu)$ with set of global minima $X^*$ if for any $\mu>0$ s.t. $f\in \mathcal{C}^-(\mu)$, $\forall \, \varepsilon>0, \exists \, \eta>0$ s.t. $\forall \, h\in \mathcal{F}_{X^*},\text{if } \|h\|_* \leq \eta$, then $f+h\in\mathcal{C}^-(\mu-\varepsilon)$.
    
    We say that $\mathcal{C}^+$ is continuous if it is continuous in all $f\in \bigcup_{L>0}\mathcal{C}^+(L)$ that admits a convex set of global minima, and $\mathcal{C}^-$ is continuous if it is continuous in all $f\in \bigcup_{\mu>0} \mathcal{C}^-(\mu)$ that admits a convex set of global minima.
\end{Def}

Note that this definition is independent from the standard notion of continuity, as we only allow $f$ to be approximated by functions in $f+\mathcal{F}_{X^*}$.

Based on the observations of Theorem \ref{continuity1} and Corollary \ref{continuity3}, if we tune a continuous FOA based on a condition $\mathcal{C}^\pm$, it is desirable for $\mathcal{C}^\pm$ to be continuous in the sense we just introduced. However, the standard properties of smoothness and strong convexity fail to be continuous:
\begin{Th} \label{continuity4}
    For any $f$  $\bar{\mu}$-strongly convex and $\bar{L}$-smooth with a set of global minima $X^*\subseteq \mathbb{R}^d$, there exists a family $\{h_\varepsilon\}_{\varepsilon>0}$ in $\mathcal{F}_{X^*}$ such that $\underset{\varepsilon \rightarrow 0}{\lim} \|h_\varepsilon\|_*=0$ and $\forall \, L,\mu>0$, there is $\varepsilon_{L,\mu}$ such that $\forall \, \varepsilon \leq \varepsilon_{L,\mu}$, $f_\varepsilon=f+h_\varepsilon$ is not $L$-smooth and not $\mu$-strongly convex.
\end{Th}

Not only smoothness and strong convexity are continuous nowhere, but also the discontinuity is not bounded: given any objective function $f$, it is possible to approximate it by a family of perturbed functions  $\{f_\varepsilon \}_{\varepsilon > 0}$ with arbitrarily bad conditioning. In particular, the explicitly construction of $\{f_\varepsilon\}_{\varepsilon >0}$ is given in the proof. Therefore, the main consequence of Theorem \ref{continuity4} is that tunings that rely on smoothness and strong convexity lack robustness.

    \section{Alternative conditioning}\label{sec:notations}
    Motivated by the weakness of strong convexity and smoothness detailed in Subsection \ref{fail_tuning} and in Theorem \ref{continuity4}, we propose here known alternative conditions that could be used to tune FOA.

Let $f\in C^1\left( \mathbb{R}^d \right)$ with convex set of minimizer $X^*$. We recall that any $x\in\mathbb{R}^d$ has an unique projection $x^*_p \in X^*$ on $X^*$: $\|x-x^*_p\|_2 = d(x,X^*)$.

\begin{Def}[Lower conditions]
    \label{lower-bounds-properties}
    Let $\mu>0$. We define:
    \begin{itemize}
        \item \emph{(Strong convexity)} $f\in \SC^-(\mu)$ iff $f(y) \geq f(x) +  \langle \nabla f(x) , y - x \rangle  + \frac{\mu}{2}\left\| y - x \right\|^2_2$, $\forall \, x, y \in \mathbb{R}^d$.
        \item \emph{(Star strong convexity)} $f\in  ~^*\!\SC^-(\mu)$ iff $f^* \geq f(x) + \langle \nabla f(x),   x^*_p - x\rangle  + \frac{\mu}{2}\left\| x^*_p - x \right\|^2_2$, $\forall \, x \in \mathbb{R}^d$.
        \item \emph{(Lower restricted secant inequality)} $f\in \RSI^-(\mu)$ iff  $ \langle \nabla f(x) , x-x^*_p\rangle  \geq \mu \left\| x-x^*_p \right\|^2_2$, $\forall \, x \in \R^d$.
        \item \emph{(Lower error bound)} $f\in \EB^-(\mu)$ iff $\left\| \nabla f(x) \right\|_2 \geq \mu\le\|x-x_p^*\ri\|_2$, $\forall \, x\in \R^d$.
        \item \emph{(Lower Polyak-{\L}ojasiewicz)} $f\in \PL^-(\mu)$ iff  $\frac{1}{2} \left\| \nabla f(x) \right\|^2_2 \geq \mu \left( f(x) - f^* \right)$, $\forall \, x\in\R^d$.
        \item \emph{(Lower quadratic growth)} $f\in \QG^-(\mu)$ iff  $f(x)-f^* \geq \frac{\mu}{2} \le\|x-x_p^*\ri\|_2^2$, $ \forall \, x\in \R^d$.
    \end{itemize}
\end{Def}

\begin{Rem}
    \label{rmk:mu0}
    A function in $\SC^-(0)$ is called \textit{convex} and a function in $^*\!\SC^-(0)$ is called \textit{star-convex}.
    Additionally, if the inequality in the definition of  $\SC^-(0)$ is strict, then  $f$ is \textit{strictly convex}.
\end{Rem}

\begin{Def}[Upper conditions]
    \label{upper-bounds-properties}
    Let $L>0$. We define: 
    \begin{itemize}
        \item \emph{(Smoothness)} $f\in \SC^+(L)$ iff  $f(y) \leq f(x) + \langle  \nabla f(x) , y - x \rangle  + \frac{L}{2}\left\| y - x \right\|_2^2$, $\forall\, x, y \in \mathbb{R}^d$.
        \item \emph{(Star smoothness)} $f\in ~^*\!\SC^+(L)$ iff   $f^* \leq f(x) + \langle \nabla f(x) , x^*_p - x \rangle + \frac{L}{2}\left\| x^*_p - x \right\|^2_2$, $ \forall\, x \in \mathbb{R}^d$.
        \item \emph{(Upper restricted secant inequality)} $f\in \RSI^+(L)$ iff  $\langle \nabla f(x) , x-x^*_p\rangle  \leq L \left\| x-x^*_p \right\|^2_2$, $ \forall \, x \in \R^d$.
        \item \emph{(Upper error bound)} $f\in \EB^+(L)$ iff $\left\| \nabla f(x) \right\|_2 \leq L \left\| x-x^*_p \right\|_2$,  $ \forall \, x\in \R^d$.
        \item \emph{(Upper Polyak-{\L}ojasiewicz)} $f\in \PL^+(L)$ iff $\frac{1}{2} \left\| \nabla f(x) \right\|^2_2 \leq L \left( f(x) - f^* \right)$,  $ \forall\, x\in \R^d$.
        \item \emph{(Upper quadratic growth)} $f\in \QG^+(L)$ iff  $f(x)-f^* \leq \frac{L}{2} \left\| x - x^*_p \right\|^2_2$, $\forall\, x\in \R^d$.
    \end{itemize}
\end{Def}

\begin{Rem}
The proposed upper and lower conditions all coincide on quadratics, with optimal $L$ and $\mu$ equal to the highest and lowest eigenvalues of the Hessian, respectively. 
\end{Rem}

\input{sections/conditions_graphs_AISTATS}

The upper and lower conditions above are related according to the graphs in \figurename \ \ref{assumptions_graph} (see proofs in Appendices \ref{App_graph_lower} and \ref{App_graph_upper}). If an implication changes the value of the constant, it is specified on the corresponding arrow. Some of the implications only hold under extended notions of $^*\!\SC^-(\mu)$ and $\SC^-(\mu)$, where $\mu$ is allowed to be negative (red arrows in \figurename \ \ref{assumptions_graph}). These notions are weaker than star convexity and convexity, respectively. 
Finally some implications are made under an additional $\QG^+$ or $\QG^-$ assumption (green arrows in \figurename \ \ref{assumptions_graph}).

In \citep{DBLP:journals/corr/KarimiNS16}, the authors already presented connections between the lower conditions, but under the assumption of global smoothness ($\SC^+(L)$) and without giving the conversion of constants. To the best of our knowledge, there is no study of the implications between upper conditions in the literature.

Theorem \ref{continuity4} showed smoothness and strong convexity are not continuous in the sense of Definition \ref{condition_continuity}.
On the other hand, the above alternatives are continuous conditions, therefore they are robust to the type of perturbations introduced in Section \ref{subsec:starnorm}:


\begin{Th}
    \label{cond_continuity1} The lower conditions
    $^*\!\SC^-$, $\RSI^-$, $\EB^-$, $\QG^-$, $\PL^-$ are continuous. 
    The upper conditions $^*\!\SC^+$, $\RSI^+$, $\EB^+$, $\QG^+$, $\PL^+$ are continuous in all functions $f\in \QG^-(\mu)$, for some $\mu>0$.
\end{Th}
\emph{Proof.} See Appendix \ref{App_cond_continuity}.

Note that since $\SC^-(\mu), ~^*\!\SC^-(\mu), \PL^-(\mu), \RSI^-(\mu),  \EB^-(\mu) \subset \QG^-(\frac{\mu^2}{L})$ (see  \figurename\ \ref{assumptions_graph}), $^*\!\SC^+$, $\RSI^+$, $\EB^+$, $\QG^+$, $\PL^+$ are continuous in any function that verifies one of the proposed lower conditions.

    \section{Gradient descent convergence}\label{sec:GD}
     \begin{table*}
     \caption{Linear rates for the GD algorithm for each pair of conditions, as function of $\kappa=\frac{L}{\mu}$. Rates marked with $^*$ hold under the additional assumption of star-convexity, while rates marked with $^\dagger$ hold under the additional assumption of convexity. Rates are colored in green if corresponding to a continuous pair of conditions and red otherwise.}

\makebox[\textwidth][c]{

\begin{tabular}{|c||c|c|c|c|c|c|}
\hline
{\color[HTML]{333333} Rates of cv} & {\color[HTML]{333333} $\SC^-(\mu)$}                                      & {\color[HTML]{333333} $^*\!\SC^-(\mu)$}               & {\color[HTML]{333333} $\PL^-(\mu)$}                  & {\color[HTML]{333333} $\RSI^-(\mu)$}                & {\color[HTML]{333333} $\EB^-(\mu)$}                  & {\color[HTML]{333333} $\QG^-(\mu)$}                  \\ \hhline{|=|=|=|=|=|=|=|}
{\color[HTML]{333333} $\SC^+(L)$}   & {\color{red} $\left( \frac{\kappa - 1}{\kappa + 1} \right)^2$} & {\color{red} $1 - \frac{1}{\kappa}$}      & {\color{red} $1 - \frac{1}{\kappa}$}       & {\color{red} $1 - \frac{1}{\kappa^2}$ / $1 - \frac{1}{2\kappa}$ *}    & {\color{red} $1 - \frac{1}{\kappa^2}$}     & {\color{red} $1 - \frac{1}{4\kappa}$ *}  \\ \hline
{\color[HTML]{333333} $\PL^+(L)$}   & {\color{red} $\left( 1 - \frac{1}{\kappa} \right)^2$}          & {\color[HTML]{008000} $1 - \frac{1}{\kappa}$}    & {\color[HTML]{008000} $1 - \frac{1}{4\kappa}$ *}  & {\color[HTML]{008000} $1 - \frac{1}{\kappa^2}$ / $1 - \frac{1}{2\kappa}$ *}    & {\color[HTML]{008000} $1 - \frac{1}{4\kappa^2}$ *}  & {\color[HTML]{008000} $1 - \frac{1}{4\kappa}$ *}  \\ \hline
{\color[HTML]{333333} $\EB^+(L)$}   & {\color{red} $\left( 1 - \frac{1}{\kappa} \right)^2$}          & {\color[HTML]{008000} $1 - \frac{1}{\kappa^2}$}    & {\color[HTML]{008000} $1 - \frac{1}{4\kappa^2}$ *}  & {\color[HTML]{008000} $1 - \frac{1}{\kappa^2}$}    & {\color[HTML]{008000} $1 - \frac{1}{4\kappa^4}$ *}  & {\color[HTML]{008000} $1 - \frac{1}{4\kappa^2}$ *}  \\ \hline
{\color[HTML]{333333} $^*\!\SC^+(L)$} & {\color{red} $\left( 1 - \frac{1}{\kappa} \right)^2$}          & {\color[HTML]{008000} $1 - \frac{1}{\kappa^2}$ $^\dagger$}  & {\color[HTML]{008000} $1 - \frac{1}{4\kappa^2}$ $^\dagger$}  & {\color[HTML]{008000} $1 - \frac{1}{\kappa^2}$ $^\dagger$}  & {\color[HTML]{008000} $1 - \frac{1}{4\kappa^4}$ $^\dagger$}  & {\color[HTML]{008000} $1 - \frac{1}{4\kappa^2}$ $^\dagger$}  \\ \hline
{\color[HTML]{333333} $\RSI^+(L)$}  & {\color{red} $\left( 1 - \frac{1}{\kappa} \right)^2$}          & {\color[HTML]{008000} $1 - \frac{1}{4\kappa^2}$ $^\dagger$} & {\color[HTML]{008000} $1 - \frac{1}{16\kappa^2}$ $^\dagger$} & {\color[HTML]{008000} $1 - \frac{1}{4\kappa^2}$ $^\dagger$} & {\color[HTML]{008000} $1 - \frac{1}{16 \kappa^4}$ $^\dagger$} & {\color[HTML]{008000} $1 - \frac{1}{16\kappa^2}$ $^\dagger$} \\ \hline
{\color[HTML]{333333} $\QG^+(L)$}   & {\color{red} $\left( 1 - \frac{1}{\kappa} \right)^2$}          & {\color[HTML]{008000} $1 - \frac{1}{4\kappa^2}$ $^\dagger$} & {\color[HTML]{008000} $1 - \frac{1}{16\kappa^2}$ $^\dagger$} & {\color[HTML]{008000} $1 - \frac{1}{4\kappa^2}$ $^\dagger$} & {\color[HTML]{008000} $1 - \frac{1}{16\kappa^4}$ $^\dagger$} & {\color[HTML]{008000} $1 - \frac{1}{16\kappa^2}$ $^\dagger$} \\ \hline
\end{tabular}
}

     \label{convergence_table}
 \end{table*}

To give some insights on the strengths of the listed conditions, we collected in Table \ref{convergence_table} the guaranteed linear convergence rates of $f(x_n)-f^*$ of the GD algorithm with constant step size and proper tuning, obtained for each pair of upper/lower conditions $f\in \mathcal{C}^+(L)\cap \mathcal{C}^-(\mu)$, as function of the condition number $\kappa=\frac{L}{\mu}$. The conditions are ordered from the strongest to the weakest, when applicable. The rates that are marked with an asterisk or a $\dagger$ symbol are only guaranteed under an additional assumption of convexity or star convexity, respectively. 
Many of these rates do not exist in the literature to the best of our knowledge. In particular, the rates
under $\QG^+(L)\cap \SC^-(\mu)$, and all the rates inherited from the known ones, as in Figure \ref{assumptions_graph}, are novel.
The rate under $\PL^+\cap ^*\!\SC^-(\mu)$ is a particular case of Theorem 3.1 in \citep{gower2019sgd} applied to the deterministic case.
For the sake of completeness, we reported rates under additional convexity assumption, although convexity suffers from the same continuity issue as strong convexity and smoothness. 

We refer to Appendix \ref{App_rates_conv} for the proofs; the exact value of the step size for the convergence of GD under each pair of upper/lower conditions is also given.

Some care needs to be taken when comparing the $\kappa$'s from different entries of the table, as the quantities involved ($L$ and $\mu$) differ according to the upper/lower conditions considered.
Notice that the condition $\PL^+(L)$ paired with any lower condition shows a convergence rate with the same dependence in $\kappa$ as $\SC^+(L)$, with the added bonus that $\PL^+(L)$ is continuous. 
Additionally, the pair $\EB^+(L) \cap \RSI^-(\mu)$ has a linear rate that depends quadratically in $\kappa$, however, this pair of conditions is weaker than other pairs ($\PL^+(L)$, $\PL^-(\mu)$, $^*\!\SC^-(\mu)$), therefore the condition number $\kappa$ for this case might be drastically smaller and it may yield a better convergence rate. Thus, these two pairs $\PL^+(L)\cap \mathcal{C}^-(\mu)$ and $\EB^+(L)\cap\RSI^-(\mu)$  look particularly promising for effectively tuning the step size of the GD algorithm.


We recall that quadratics give a lower bound $\le(\frac{\kappa - 1}{\kappa + 1} \ri)^2$ for the convergence rate for GD with fixed step size. Since all the conditions listed in this paper coincide on quadratics, such a lower bound applies to any pair of upper/lower conditions. However, it may be not tight for some pairs of conditions.

Finally, we complete Table \ref{convergence_table} by mentioning the sublinear convergence speed we have under any upper condition and convexity or star-convexity ($\mathcal{C}^+(L)\cap \SC^-(0)$ or $\mathcal{C}^+(L)\cap \,^*\!\SC^-(0)$).
While it is known that GD has a rate of convergence of order $\mathcal{O}\left(\frac{1}{n}\right)$ if $f \in \SC^+(L)\cap \SC^-(0)$ (see e.g. \citep{bansal2017potential}), the same rate can be achieved under $\PL^+(L)\cap \, ^*\!\SC^-(0)$ for the best iterate (or the average under convexity).
For a complete proof, we refer to Appendix \ref{App_rates_conv}.

In \citep{ghadimi2015global}, the authors prove the same rate of convergence under convexity and $\QG^+(L)$ for the average iterate. They also obtain the same rate under those very weak conditions for the last iterate using an extra momentum term (following the heavy ball procedure).

    \section{Discussion}\label{sec:applications} 

In this section we discuss how the use of alternative conditions impacts our understanding of the behavior of Polyak's Heavy-Ball method \citep{Polyak}:
\begin{equation}
    x_{n+1} = x_n - \alpha\nabla f(x_n) + \beta(x_n-x_{n-1})
\end{equation}
where the step size $\alpha$ and the momentum $\beta$ are the hyperparameters. It is well known that the optimal hyperparameters of HB on a strongly convex \emph{quadratic} function with minimum eigenvalue $\mu$ and maximum eigenvalue $L$ of the Hessian are:
\begin{equation}
\label{HB1}
\alpha = \frac{4}{(\sqrt{L}+\sqrt{\mu})^2},\qquad \beta = \left(\frac{\sqrt{\kappa}-1}{\sqrt{\kappa}+1}\right)^2 
\end{equation}
with $\kappa = \frac{L}{\mu}$.  This tuning yields a linear convergence rate for $f(x_n)-f^*$ equal to $\le(\frac{\sqrt{\kappa}-1}{\sqrt{\kappa}+1}\ri)^2$, which approaches the lower bound for $L$-smooth and $\mu$-strongly convex functions \citep{bubeck2014convex}.

In \citep{Lessard} the authors introduced the following piecewise quadratic function: $f_{\rm LRP}\in C^1(\R)$ with derivative
\begin{equation}
     f'_{\rm LRP}(x) =
    \begin{cases}
    25x & x<1\\
    x + 24 & 1\leq x \leq 2\\
    25x - 24 & x>2
    \end{cases}
\end{equation}
The authors showed that for the initial value $x_0=3.3$ and using the same tuning rule \eqref{HB1} but with the $L$ of smoothness and the $\mu$ of strong convexity, HB does not converge. 

While applying the tuning rule \eqref{HB1} to non-quadratic functions itself is arbitrary, the choice of smoothness and strong convexity values as generalizations of the maximum and minimum eigenvalues is particularly problematic: all conditions introduced in Section \ref{sec:notations} coincide on quadratic functions and there is no strong evidence to prefer $\SC^-(\mu)$ and $\SC^+(L)$ as tuning conditions.

Since this function has been used as an example of inconsistent behavior from HB, it is natural to question how using continuous conditions as generalizations of the biggest and smallest eigenvalues of a quadratic function may affect the convergence.

All the upper conditions on $f_{\rm LRP}$ give the same parameter $L=25$, while the lower conditions give $\mu_{\SC^-}=1$, $\mu_{^*\!\SC^-}=7$, $\mu_{\RSI^-}=\mu_{EB^-}=13$, $\mu_{\PL^-}=\frac{169}{19}$ and $\mu_{\QG^-}=19$.

\begin{figure}
\centering
    \includegraphics[scale=0.45]{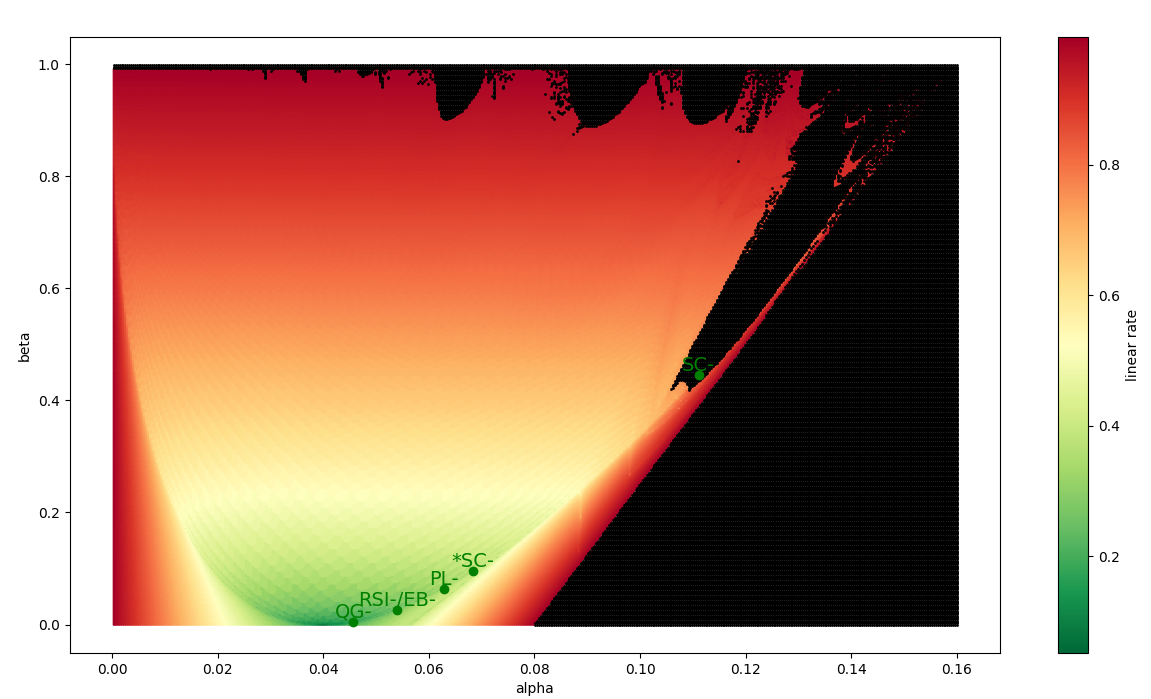}
    \caption{Convergence rate of HB on $f_{\rm LRP}$ with starting point $x_0 = 3.3$ for different tunings of $\alpha$, $\beta$. Black areas mean no linear convergence}
    \label{fig:HB}
\end{figure}

In \figurename \ \ref{fig:HB} we present the linear convergence rates experimentally obtained for 200,000 values of $(\alpha,\beta)$. We also indicate the hyperparameters corresponding to tuning rule \eqref{HB1} using different lower conditions. We immediately observe that while the generalization based on strong convexity falls into the black region (no linear convergence), other conditions all offer excellent convergence properties. This might suggest that the divergent behavior is caused by relying on strong convexity for tuning rather than by the algorithm itself, and highlights how weaker conditions are essential to understand FOA behaviors, even on functions that verify smoothness and strong convexity.

    \section{Conclusion}\label{sec:conclusion}
    In this paper we presented an argument on the necessity to adopt different conditions from the ones classically used (smoothness and strong convexity), in order to tune the hyperparameters of FOA in a meaningful way. 
Via a new notion of continuity of a condition number, we have established that the properties of strong convexity and smoothness have an important weakness resulting in a lack of robustness for first-order algorithms tuned on them. We have presented promising alternatives that do not share this weakness and given examples of the benefits of a theoretical framework based on these conditions.  We have proposed an extensive study of the relationships between these conditions and provided their guaranteed convergence rates for GD.

The study of the convergence properties of optimization algorithms largely depends on their tuning, hence understanding its underlying conditions leads to a better comparison between them and improves the algorithm performances, as illustrated in Section \ref{sec:applications}.


While it is well known that some optimization algorithms (e.g. Nesterov Accelerated Gradient, \citep{Nesterov}) can approach the lower bound of convergence rates achievable for $\mu$-strongly convex and $L$-smooth functions, as function of $\kappa = \frac{L}{\mu}$, lower bounds based on alternative condition numbers will result in different optimality results, which we leave to future work.


    
    
    
    \paragraph{Acknowledgements.}
        We thank Nicolas Loizou for his valuable advices. We also thank Leonard Boussioux and Aymeric Dieuleveut for their proofreading and useful feedbacks.
        
        This work was partially supported by the FRQNT new researcher program (2019-NC-257943), the NSERC Discovery grant (RGPIN-2019-06512), a startup grant by IVADO, a Microsoft Research collaborative grant and a Canada CIFAR AI chair.
    
    \bibstyle{plain}
    \bibliography{references_arxiv}~\nocite{*}
    
    
    \appendix

    \section{Continuity} \label{App_continuity}
    \paragraph{Proof of Theorem \ref{continuity1}}
We will prove the theorem by induction. Let $f\in C^1(\R^d)$ with a set of global minima $X^*$. For $h\in\mathcal{F}_{X^*}$, let $g=f+h\in f+\mathcal{F}_{X^*}$. For $n=0$,   $x_0(\cdot,\mathcal{A}_\theta,x_0)\equiv x_0$ is clearly continuous (in the sense of $\le\| \cdot \ri\|_*$) for any fixed initial point $x_0 \in \R^d$; assume that the continuity property is verified up to some $n\in\mathbb{N}$: i.e. $\forall \, \epsilon >0$ $\forall \, i = 0,\ldots, n$, $\exists \, \eta= \eta(\epsilon, i, \mathcal{K})>0$ such that for $g \in f+\mathcal{F}_{X^*}$, if $\|f-g\|_*<\eta$, then $\forall \, x_0\in\mathcal{K}$, $\|x_i(f, \mathcal{A}_\theta, x_0) - x_i(g,\mathcal{A}_\theta, x_0) \|_2<\epsilon$.

 Let $\epsilon>0$ and 
 $$x_{n+1}(f,\mathcal{A}_\theta,x_0)=\mathcal{A}_\theta\Big( \{x_{i}\}_{i=0\ldots n}, \{f(x_i)\}_{i=0\ldots n}, \{\nabla f(x_i)\}_{i=0\ldots n} \Big);$$
$\mathcal{A}_\theta$ being a continuous FOA implies that given $\epsilon>0$, there exists $\delta>0$ such that if $\forall \, i=0,\ldots, n$
\begin{align}
& \le\|x_{i}(f,\mathcal{A}_\theta,x_0)-x_{i}(g,\mathcal{A}_\theta,x_0)\ri\|_2 < \delta \label{ineq1}\\
&\le\|f(x_i(f,\mathcal{A}_\theta,x_0))-g(x_i(g,\mathcal{A}_\theta,x_0))\ri\|_2< \delta \label{ineq2}\\
&\le\|\nabla f(x_i(f,\mathcal{A}_\theta,x_0))-\nabla g(x_i(g,\mathcal{A}_\theta,x_0))\ri\|_2< \delta \label{ineq3}
\end{align}
for $f\in C^1(\mathbb{R}^d)$, $g\in f+\mathcal{F}_{X^*}$,  then the claim follows
$$\|x_{n+1}(f,\mathcal{A}_\theta,x_0)-x_{n+1}(g, \mathcal{A}_\theta, x_0)\|_2 < \epsilon.$$

The idea now is to quantify how "close" $f$ and $g$ need to be (in $\le\|\cdot\ri\|_*$-norm) in order to ensure that the above inequalities are satisfied.

\underline{For equation \eqref{ineq1}}:  by recurrence hypothesis ($n\in \mathbb{N}$ is finite), given $\delta >0$ there exists $ \eta_1 = \eta_1(\delta)$ (simply consider $ =  \min_{i=1,\ldots,n} \{\eta(\delta, i, \mathcal{K}) \}>0$) such that $\forall \, g \in f+\mathcal{F}_{X^*}, \|f-g\|_*<  \eta_1$, then $\forall \, x_0\in\mathcal{K}$, $\|x_i(f,\mathcal{A}_\theta,x_0)-x_i(g,\mathcal{A}_\theta,x_0)\|_2 < \delta$, $ \forall \, i =1,\ldots,  n$.

\underline{For equation \eqref{ineq2}}:
\begin{gather}
\|\nabla f(x_i(f,\mathcal{A}_\theta,x_0))-\nabla g(x_i(g,\mathcal{A}_\theta,x_0))\|_2 \nonumber \\
\leq \|\nabla f(x_i(f,\mathcal{A}_\theta,x_0))-\nabla f(x_i(g,\mathcal{A}_\theta,x_0))\|_2 
+ \|\nabla f(x_i(g,\mathcal{A}_\theta,x_0))-\nabla g(x_i(g, \mathcal{A}_\theta, x_0))\|_2
\end{gather}

The first term can be easily estimated: since $f\in C^1(\mathbb{R}^d)$, given $\delta>0$ there exists $\rho = \rho(\delta)>0$ such that $\forall \, x,y \in \R^d$ with $\|x-y\|_2< \rho$, then $ \|\nabla f(x)-\nabla f(y)\|_2< \frac{\delta}{2}$ and $ \| f(x)- f(y)\|_2< \frac{\delta}{2}$.  In particular, for such a $\rho>0$, $\exists \, \eta_2 = \eta_2(\rho, \delta)>0$ such that for $\|f-g\|_* < \min\{\eta_1, \eta_2\}$, then $ \|x_i(f,\mathcal{A}_\theta,x_0)-x_i(g,\mathcal{A}_\theta,x_0)\|_2 < \rho$  $\forall \, i=0,\ldots,n$.
Therefore,
\begin{align}
\|\nabla f(x_i(f,\mathcal{A}_\theta,x_0))-\nabla f(x_i(g, \mathcal{A}_\theta, x_0))\|_2< \frac{\delta}{2} \qquad \forall \, i=0,\ldots,n. \label{eq16}
\end{align}

Regarding the second term, we first introduce the quantity 
$$R_{f,n}:=\underset{i=0,\ldots,n}{\max}\le\{\underset{x_0\in\mathcal{K}}{\sup}\  d\le(x_i(f,\mathcal{A}_\theta,x_0),X^*\ri)\ri\} $$ 
$x_i(f,\mathcal{A}_\theta,\cdot)$ is a finite composition of continuous functions and is therefore continuous (in $x_0$). This ensures that the image of $\mathcal{K}$ is a compact and thus that $R_{f,n}$ is indeed finite.
\begin{gather}
\|\nabla f(x_i(g,\mathcal{A}_\theta,x_0))-\nabla g(x_i(g, \mathcal{A}_\theta, x_0))\|_2 \nonumber \\
=\frac{\|\nabla f(x_i(g,\mathcal{A}_\theta,x_0))-\nabla g(x_i(g, \mathcal{A}_\theta, x_0))\|_2}{d\le(x_i(g,\mathcal{A}_\theta,x_0),X^*\ri)}d\le(x_i(g,\mathcal{A}_\theta,x_0),X^*\ri)
\leq \|f-g\|_* \,R_{g,n} 
\end{gather}
We want to claim that if $\|f-g\|_*$ is small enough (for $g \in f+\mathcal{F}_{X^*}$), then $R_{g,n} < R_{f,n} + \delta$: indeed, if $g\in f+ \mathcal{F}_{X^*}$ is such that $\|f-g\|_*< \min \{ \eta_1, \eta_2 \}$, by recurrence hypothesis we have $\forall \, i=0,\ldots, n$ 
\begin{align}
d\le(x_i(g, \mathcal{A}_\theta,x_0),X^*\ri) & = \inf_{x^*\in X^*} \|x_i(g, \mathcal{A}_\theta,x_0)-x^*\|_2 \nonumber \\
& \leq \inf_{x^*\in X^*} \le\{ \|x_i(g, \mathcal{A}_\theta,x_0)-x_i(f,\mathcal{A}_\theta,x_0)\|_2+\|x_i(f, \mathcal{A}_\theta,x_0)-x^*\|_2 \ri\} \nonumber \\
& = \|x_i(g, \mathcal{A}_\theta,x_0)-x_i(f,\mathcal{A}_\theta,x_0)\|_2+  \inf_{x^*\in X^*} \|x_i(f, \mathcal{A}_\theta,x_0)-x^*\|_2  \nonumber \\
& = \|x_i(g, \mathcal{A}_\theta,x_0)-x_i(f,\mathcal{A}_\theta,x_0)\|_2+  d\le(x_i(f, \mathcal{A}_\theta,x_0),X^*\ri) \nonumber \\
&< \delta + R_{f,n}. 
\end{align}

Then, 
\begin{gather}
\|\nabla f(x_i(g,\mathcal{A}_\theta,x_0))-\nabla g(x_i(g, \mathcal{A}_\theta, x_0))\|_2 \leq \|f-g\|_*(R_{f,n}+\delta) < \frac{\delta}{2}
\end{gather}
as long as $\|f-g\|_* < \min\{\eta_1, \eta_2, \frac{\delta}{2(R_{f,n}+\delta)}\}$.

In conclusion,  $\forall \, i=0,\ldots, n$
\begin{gather}
\label{cont1}
\|\nabla f(x_i(f,\mathcal{A}_\theta,x_0))-\nabla g(x_i(g,\mathcal{A}_\theta,x_0))\|_2 \nonumber \\
\leq \|\nabla f(x_i(f,\mathcal{A}_\theta,x_0))-\nabla f(x_i(g,\mathcal{A}_\theta,x_0))\|_2 
+ \|\nabla f(x_i(g,\mathcal{A}_\theta,x_0))-\nabla g(x_i(g, \mathcal{A}_\theta, x_0))\|_2 \nonumber \\
\leq \frac{\delta}{2} + \frac{\delta}{2} = \delta
\end{gather}

\underline{For equation \eqref{ineq3}}: similarly, we have
\begin{gather}
\|f(x_i(f,\mathcal{A}_\theta,x_0))- g(x_i(g,\mathcal{A}_\theta,x_0))\|_2 \nonumber \\
\leq \|f(x_i(f,\mathcal{A}_\theta,x_0))- f(x_i(g,\mathcal{A}_\theta,x_0))\|_2 + \|f(x_i(g,\mathcal{A}_\theta,x_0))-g(x_i(g, \mathcal{A}_\theta, x_0))\|_2
\end{gather}

The first term is bounded by $\delta/2$ thanks the same argument as in \eqref{eq16}. The second term is bounded in the following way: call $\bar x = x_i(g,\mathcal{A}_\theta,x_0)$ and let $\bar x^*_p \in X^*$ the projection of $\bar x$ on $X^*$. Note that $ \forall \, t\in[0,1]$, we have $d\le(  \bar x^*_p +t(\bar x- \bar x^*_p),X^*\ri) \leq t\le\|\bar x- \bar x^*_p\ri\|_2$ since $\bar x_p^*\in X^*$.
It follows that
\begin{align}
 |f(\bar x)-g(\bar x)| & = |f(\bar x)-g(\bar x) -( f^* - g^* )| \nonumber \\
 &=\le|\int_{0}^{1} \langle \nabla(f-g)(\bar x^*_p+t(\bar x-\bar x^*_p)) ,  \bar x^*_p - \bar x\rangle \ \d t \ri|\nonumber\\
  &\leq\int_{0}^{1} \le\| \nabla(f-g)( \bar x^*_p+t(\bar x-\bar x^*_p)) \ri\|_2 \le\|  \bar x^*_p - \bar x  \ri\|_2 \ \d t\nonumber\\
& = \int_{0}^{1} \dfrac{\le\| \nabla(f-g)( \bar x^*_p+t(\bar x-\bar x^*_p)) \ri\|_2}{d\le( \bar x^*_p+t(\bar x-\bar x^*_p), X^*  \ri)}\,  d\le( \bar x^*_p+t(\bar x-\bar x^*_p), X^*  \ri) \le\| \bar x^*_p - \bar x  \ri\|_2 \ \d t\nonumber\\
& \leq \int_{0}^{1} \dfrac{\le\| \nabla(f-g)( \bar x^*_p+t(\bar x-\bar x^*_p)) \ri\|_2}{d\le( \bar x^*_p+t(\bar x-\bar x^*_p), X^*  \ri)} t \le\|  \bar x^*_p - \bar x  \ri\|_2^2 \ \d t\nonumber\\
 &\leq \|f-g\|_* d\le(   \bar x, X^*  \ri)^2 \int_{0}^{1}t \ \d t\nonumber \\
 &\leq \|f-g\|_* \frac{(R_{f,n}+\delta)^2}{2}\nonumber\\
 &< \frac{\delta}{2}
\end{align}
as long as $\|f-g\|_* < \min\le\{\eta_1, \eta_2, \frac{\delta}{2(R_{f,n}+\delta)}, \frac{\delta}{(R_{f,n} +\delta)^2}\ri\}$.

In conclusion, $\forall \, i=0,\ldots, n$
\begin{gather}
\label{cont3}
\|f(x_i(f,\mathcal{A}_\theta,x_0))- g(x_i(g,\mathcal{A}_\theta,x_0))\|_2 \nonumber \\
\leq \|f(x_i(f,\mathcal{A}_\theta,x_0))- f(x_i(g,\mathcal{A}_\theta,x_0))\|_2 
+ \|f(x_i(g,\mathcal{A}_\theta,x_0))-g(x_i(g, \mathcal{A}_\theta, x_0))\|_2 \nonumber \\
< \frac{\delta}{2} + \frac{\delta}{2} = \delta.
\end{gather}
\qed

\paragraph{Proof of Corollary \ref{continuity3}}

Let $N_{\mathcal{K}}={\sup}_{x_0\in\mathcal{K}}N_{x_0}$ (note that $N_{\mathcal{K}}<+\infty$, since $\mathcal{K}$ is compact). We will note $g=f+h$ for $h\in\mathcal{F}_{X^*}$.


For $x_0\in\mathcal{K}$, we have \begin{gather*}
    x_{N_{x_0}-1}(f,\mathcal{A}_\theta,x_0)\notin \mathcal{B}(X^*,\varepsilon)
    \quad \text{and} \quad
    x_{N_{x_0}}(f,\mathcal{A}_\theta,x_0)\in \mathcal{B}(X^*,\varepsilon)
\end{gather*}
and thanks to Theorem \ref{continuity1}, there exists $\eta>0$ such that for any $g\in f+\mathcal{F}_{X^*}$, if $||f-g||_*\leq \eta$ then $\forall\, i\leq N_\mathcal{K}$, $\forall \, x_0\in\mathcal{K}$, $\le\|x_i(f,\mathcal{A}_\theta,x_0)-x_i(g,\mathcal{A}_\theta,x_0)\ri\|_2\leq \delta$. Therefore,
\begin{gather*}
    x_{N_{x_0}-1}(g,\mathcal{A}_\theta,x_0)\notin \mathcal{B}(X^*,\varepsilon-\delta)
    \quad \text{and} \quad
    x_{N_{x_0}}(g,\mathcal{A}_\theta,x_0)\in \mathcal{B}(X^*,\varepsilon+\delta).
\end{gather*}
\qed 

\paragraph{Proof of Proposition \ref{prop-continuity}}
$f$ is a piecewise quadratic with second derivative $f''(x)=(2 + \frac{2}{\varepsilon})$ for $x\in[1,1+\varepsilon^2]$ and $f''(x)=2$ elsewhere. 
Therefore, the optimal $\mu$ of strong convexity is $2$ and the optimal $L$ of smoothness is $2+\frac{2}{\varepsilon}$: $f\in \SC^-(2)\cap \SC^+(2+\frac{2}{\varepsilon})$.

Consider the gradient descent update rule with step size $\alpha = \frac{1}{2}$:
\begin{gather*}
x - \frac{1}{2} f'(x) = \begin{cases}
0 &   x\leq 1\\
\frac{x-1}{\varepsilon} & 1\leq x\leq 1+\varepsilon^2\\
-\varepsilon & x\geq 1+\varepsilon^2
\end{cases}
\end{gather*}
It is easy to see that $|x-\frac{1}{2} f'(x)|\leq \varepsilon |x|$, which proves the linear convergence rate of $f_\varepsilon$ with tuning $\alpha=\frac{1}{2}$; in fact, for $\varepsilon\leq 1$, the algorithm can converge to $x^*=0$ in at most two steps.

Let us now assume we use the standard tuning based on strong convexity and smoothness
$$\alpha=\frac{2}{\mu_\varepsilon+L_\varepsilon}=\frac{\varepsilon}{2\varepsilon+1}.$$

We then have 
\begin{gather*}
x - \alpha f'(x) = \begin{cases}
\frac{x}{2\varepsilon+1} &   x\leq 1\\
\frac{2-x}{2\varepsilon+1} & 1\leq x\leq 1+\varepsilon^2\\
\frac{x-2\varepsilon^2}{2\varepsilon+1} & x\geq 1+\varepsilon^2
\end{cases}
\end{gather*}
which leads to $$\forall \, x\in \R,\quad |x-\alpha f'(x)|\geq \frac{1-\varepsilon^2}{(2\varepsilon+1)(1+\varepsilon^2)}|x|.$$
\qed

\paragraph{Proof of Theorem \ref{continuity4}}

Let $f$ a $\bar{L}$-smooth and $\bar{\mu}$-strongly convex function with a set of minima $X^*\subseteq \mathbb{R}^d$. Note that strong convexity implies $X^*=\{x^*\}$.  

Let $\varepsilon>0$. We define the function $\omega_\varepsilon\in C^1(\mathbb{R})$ by $\omega_\varepsilon(0)=0$ and its derivative:
\begin{gather*}
\omega_\varepsilon'(t) = \begin{cases}
0 & t\leq 1-\varepsilon^2\\
\frac{1-t-\varepsilon^2}{\varepsilon}& 1-\varepsilon^2\leq t \leq 1\\
\frac{t-\varepsilon^2-1}{\varepsilon} & 1\leq t\leq 1+\varepsilon^2\\
0 &  1+\varepsilon^2 \leq t
\end{cases}
\end{gather*}
It is easy to see that  $|\omega_\varepsilon'(t)|\leq \varepsilon$, $\forall \, t \in \R$.

Let $z\in\mathbb{R}^d\setminus\{x^*\}$ and define
\begin{align}
&\phi(x):=\frac{\langle x - x^* , z - x^*\rangle}{\le\|z-x^*\ri\|_2}, \qquad x\in\mathbb{R}^d\\
&f_\varepsilon(x):=f(x) + \omega_\varepsilon \circ \phi(x)
\end{align}
Note that $\omega_\varepsilon\circ\phi(x^*)=0$ and
\begin{gather}
\nabla (f-f_\varepsilon)(x)=\nabla \le(\omega_\varepsilon\circ\phi\ri)(x)=\frac{z-x^*}{\|z-x^*\|_2}\omega_\varepsilon'\circ\phi(x);
\end{gather}
for $x\in\mathbb{R}^d$, 
\begin{align}
&\text{if } \phi(x)\leq 1-\varepsilon^2, \text{then }||\nabla(f-f_\varepsilon)(x)||_2=0\\
&\text{if } \phi(x)\geq 1-\varepsilon^2, \text{then }||\nabla(f-f_\varepsilon)(x)||_2\leq \varepsilon\leq \frac{\varepsilon}{1-\varepsilon^2}||x-x^*||_2
\end{align}
since $\phi(x)\geq 1-\varepsilon^2$ implies $\|x-x^*\|_2\geq 1-\varepsilon^2$.  Therefore, $f-f_\varepsilon\in\mathcal{F}_{X^*}$ and $\|f-f_\varepsilon\|_*\leq\frac{\varepsilon}{1-\varepsilon^2}\rightarrow 0$ when $\varepsilon\rightarrow 0$.

Let $L, \mu>0$. We now want to prove that for $\varepsilon$ sufficiently small, $f_\epsilon$ is not $L$-smooth and not $\mu$-strong convex.
Consider 
\begin{align*}
&x=x^*+(1-\varepsilon^2)\frac{z-x^*}{\|z-x^*\|_2}\\
&y=x^*+\frac{z-x^*}{\|z-x^*\|_2}
\end{align*}
so that we have $\phi(x)=1-\varepsilon^2$, $\phi(y)=1$, and $y-x=\varepsilon^2\frac{z-x^*}{\|z-x^*\|_2}$.
Since $f$ is $L_f$-smooth, $\forall \, \varepsilon>0$ we have
\begin{gather}
    f_\varepsilon(y)-f_\varepsilon(x)- \langle \nabla f_\varepsilon(x), y-x \rangle \nonumber \\
    = f(y)-f(x)- \langle \nabla f(x),y-x\rangle + \omega_\varepsilon \circ \phi(y)-\omega_\varepsilon \circ \phi(x) - \langle \nabla \omega_\varepsilon \circ \phi(x) ,y-x\rangle \nonumber\\
    \leq \frac{L_f}{2}||x-y||^2_2 + \omega_\varepsilon(1) - \omega_\varepsilon(1-\varepsilon^2) - \varepsilon^2 \omega_\varepsilon'(1-\varepsilon^2)\nonumber\\
    =\frac{L_f}{2}||x-y||^2_2 - \frac{\varepsilon^3}{2}
    = \le(\frac{L_{f}}{2} - \frac{1}{2\varepsilon}\ri)\|x-y\|_2^2;
\end{gather}
therefore, if we pick $\varepsilon$ such that $\frac{1}{\varepsilon}>L_f-\mu$, then $f_\varepsilon$ is not $\mu$-strong convex.

Similarly, consider
\begin{align*}
&x=x^*+\frac{z-x^*}{\|z-x^*\|_2}\\
&y=x^*+(1+\varepsilon^2)\frac{z-x^*}{\|z-x^*\|_2}
\end{align*}
So that we have $\phi(x)=1$, $\phi(y)=1+\varepsilon^2$, and $y-x=\varepsilon^2\frac{z-x^*}{\|z-x^*\|_2}$.
Since $f$ is $\mu_f$-strong convex, $\forall \, \varepsilon >0$ we have
\begin{gather}
    f_\varepsilon(y)-f_\varepsilon(x)- \langle \nabla f_\varepsilon(x), y-x\rangle  \nonumber \\
    = f(y)-f(x)- \langle \nabla f(x), y-x\rangle + \omega_\varepsilon \circ \phi(y)-\omega_\varepsilon \circ \phi(x) - \langle \nabla \omega_\varepsilon \circ \phi(x) , y-x\rangle \nonumber\\
    \geq \frac{\mu_f}{2}\|x-y\|^2_2 + \omega_\varepsilon(1+\varepsilon^2) - \omega_\varepsilon(1) - \varepsilon^2 \omega_\varepsilon'(1)\nonumber\\
    = \frac{\mu_f}{2}\|x-y\|^2_2 + \frac{\varepsilon^3}{2} = 
     \le(\frac{\mu_f}{2} + \frac{1}{2\varepsilon}\ri)\|x-y\|_2^2;
\end{gather}
therefore, if we pick $\varepsilon$ such that $\frac{1}{\varepsilon}>L-\mu_f$, then $f_\varepsilon$ is not $L$-smooth.\\
Finally, for any $\varepsilon\leq \min\{ \frac{1}{\max\{1,L_f-\mu\}},\frac{1}{\max\{1,L-\mu_f\}}\}$, $f_\varepsilon$ is not $L$-smooth and not $\mu$-strong convex, which concludes the proof.
\qed

    \section{Proof of Theorem \ref{cond_continuity1}}\label{App_cond_continuity}
    Note that all lower conditions listed in the theorem are continuous without any additional constraint; on the other hand, the upper conditions require the assumption of the objective function $f$ to belong to $\QG^-(\mu)$ (for some $\mu>0$) in order to be continuous. 

We stress that this extra condition is a mild adding, since tuning of a FOA usually requires $f$ to satisfy both an upper and a lower condition (and $\QG^-(\mu)$ is the weakest among the conditions we proposed). On the other hand, this is necessary to guarantee that the set of minimizer for the original $f$ and the perturbed $f+h$, $h \in \mathcal{F}_{X^*}$, are the same.

\underline{Continuity of $^*\!\SC^-$ and $^*\!\SC^+$:}
Given $f \in ~^*\!\SC^+(L)$: $f^* \leq f(x) +\langle \nabla f(x) , x^*_p-x \rangle + \frac{L}{2}\| x-x^*_p\|_2^2$, $\forall \, x\in\R^d$ (with $x^*_p\in X^*$ the corresponding projection point onto $X^*$). Consider $g = f+h$, $h \in\mathcal{F}_{X^*}$ with $\| f-g\|_* = \|h\|_* = \sup \frac{\|\nabla f(x) - \nabla g(x)\|_2}{d(x,X^*)} =  \sup \frac{\|\nabla h(x) \|_2}{d(x,X^*)} \leq \frac{\epsilon}{3}$, then 
\begin{align}
g^* = f^* \leq f(x) + \langle \nabla f(x) , x^*_p-x\rangle + \frac{L}{2}\| x-x^*_p\|_2^2,
\end{align}
where $g^*$ is the value of $g$ at each point of $X^*$.  
Note that $\forall \, x\in \R^d\setminus X^*$
\begin{align}
 \langle \nabla f(x) , x^*_p-x\rangle &=  \langle \nabla f(x) - \nabla g(x)  , x^*_p -x\rangle  + \langle \nabla g(x) , x^*_p-x\rangle  \nonumber \\
 & \leq \le\| \nabla f(x) - \nabla g(x) \ri\|_2 \le\|x - x^*_p\ri\|_2+\langle  \nabla g(x) , x^*_p-x\rangle \nonumber \\
 &\leq \|f-g\|_*  \le\|x - x^*_p\ri\|_2^2+\langle \nabla g(x) , x^*_p-x\rangle \nonumber \\
&\leq \frac{\epsilon}{3} \le\|x - x^*_p\ri\|_2^2+\langle \nabla g(x) , x^*_p-x\rangle
\end{align}
and 
\begin{align}
0= h^* &= \omega(x) + \int_0^1 \langle \nabla h(x + t(x^*_p-x)) , x^*_p-x\rangle \ \d t \nonumber \\
&\leq  h(x) + \int_0^1 \le\| \nabla h(x + t(x^*_p-x))\ri\|_2 \|x-x^*_p\|_2 \ \d t  \nonumber \\
& = h(x) + \int_0^1 \dfrac{\le\| \nabla h(x + t(x_p^*-x))\ri\|_2}{d(x + t(x^*_p-x),X^*)} \, d(x + t(x^*_p-x), X^*) \|x-x^*_p\|_2 \ \d t  \nonumber \\
&\leq  h(x) +   \le\|  h \ri\|_* \|x-x^*_p\|^2_2 \int_0^1 1-t \, \d t =  h(x) + \frac{1}{2}  \le\|  f-g \ri\|_* \|x-x^*_p\|_2^2 \nonumber \\
& \leq  h(x) +   \frac{\epsilon}{6}  \|x-x^*\|^2_2
\end{align}
where we used $ d(x + t(x^*_p-x), X^* )  = (1-t)  \|x-x^*_p \|_2$, $\forall\, t\in[0,1] $ (indeed any point lying on the line segment $x + t(x^*_p-x)$ has projection onto $X^*$ equal to $x^*_p$).

Therefore, $\forall \, x\in \R^d\setminus X^*$
\begin{align}
g^*  &\leq f(x) + \le[  \frac{\epsilon}{3} \| x-x^*_p\|^2_2 + \langle \nabla g(x) , x^*_p-x\rangle  \ri] + \frac{L}{2}\| x-x^*_p\|_2^2 \nonumber \\
&\leq f(x) + h(x)  + \frac{\epsilon}{6} \| x-x^*_p\|^2_2 + \frac{\epsilon}{3} \| x-x^*_p\|^2_2 + \langle \nabla g(x) , x^*_p-x\rangle   + \frac{L}{2}\| x-x^*\|_2^2 \nonumber \\
& = g(x) + \langle \nabla g(x) , x^*_p-x\rangle  + \frac{L+\epsilon}{2}\| x-x^*_p\|_2^2
\end{align}
(for $x=x^* \in X^*$ the inequality is trivial), i.e. $g \in ~^*\!\SC(L+\epsilon)$.

Similarly, given $f \in ~^*\!\SC^-(\mu)$: $f^* \geq f(x) +\langle \nabla f(x) , x^*_p-x\rangle + \frac{\mu}{2}\| x-x^*_p\|_2^2$, $\forall \, x\in\R^d$. Consider $g = f+h$, $h \in\mathcal{F}_{X^*}$ with $\| f-g\|_* = \|h\|_* = \sup \frac{\|\nabla f(x) - \nabla g(x)\|_2}{d(x,X^*)} =  \sup \frac{\|\nabla h(x) \|_2}{d(x,X^*)} \leq \frac{\epsilon}{3} < \mu$, then
\begin{align}
g^* = f^* \geq f(x) + \langle\nabla f(x) , x^*_p-x\rangle + \frac{\mu}{2}\| x-x^*_p\|_2^2.
\end{align}
$\forall \, x\in \R^d\setminus X^*$
\begin{align}
\langle \nabla f(x) , x^*_p-x\rangle &=  \langle \nabla f(x) - \nabla g(x),  x^*_p -x\rangle + \langle \nabla g(x) , x^*_p-x\rangle  \nonumber \\
 & \geq - \le\| \nabla f(x) - \nabla g(x) \ri\|_2 \le\|x - x^*_p\ri\|_2+ \langle \nabla g(x) , x^*_p-x \nonumber \\
 &\geq - \|f-g\|_*  \le\|x - x^*_p\ri\|_2^2+ \langle \nabla g(x) , x^*_p-x\rangle   \\
&\geq -\frac{\epsilon}{3} \le\|x - x^*_p\ri\|_2^2+\langle \nabla g(x) , x^*_p-x\rangle
\end{align}
and 
\begin{align}
0= h^* &= \omega(x) + \int_0^1 \langle \nabla h(x + t(x^*_p-x)) , x^*_p-x\rangle \ \d t \nonumber \\
&\geq  h(x) - \int_0^1 \le\| \nabla h(x + t(x^*_p-x))\ri\|_2 \|x-x^*_p\|_2 \ \d t  \nonumber \\
&\geq  h(x) - \frac{1}{2}  \le\|  f-g \ri\|_* \|x-x^*_p\|_2^2 \nonumber \\
& \geq  h(x) -   \frac{\epsilon}{6}  \|x-x^*_p\|^2_2
\end{align}

Therefore, $\forall \, x\in \R^d\setminus X^*$, $g^*   \geq g(x) + \langle \nabla g(x) , x^*-x\rangle  + \frac{\mu-\epsilon}{2}\| x-x^*\|_2^2$ and for $x=x^* \in X^*$ the inequality is trivial: $g \in ~^*\!\SC(\mu-\epsilon)$.

\underline{Continuity of $\RSI^-$ and $\RSI^+$:}
Given $f \in \RSI^+(L)$: $\forall \, x\in\R^d$,  $\langle \nabla f(x) , x-x^*_p\rangle  \leq L \le\|x-x^*_p \ri\|^2_2  $. Consider $g = f+h$ with $h \in \mathcal{F}_{X^*}$ such that $\|h\|_* = \sup_{x\in\R^d\setminus\{X^*\}} \frac{\le\| \nabla h(x) \ri\|_2 }{ d(x,X^*) } \leq \epsilon $, then we have $\forall \, x\in \R^d \setminus X^*$
\begin{align}
\langle \nabla g(x) , x-x_p^*\rangle   &= \langle \nabla f(x) + \nabla h(x) , x-x^*_p\rangle  = \langle \nabla f(x) , x-x^*_p\rangle +\langle \nabla h(x) , x-x^*_p\rangle \nonumber \\
&\leq L \le\|x-x^*_p \ri\|^2_2 + \le\| \nabla h(x)\ri\|_2 \le\|x-x_p^* \ri\|_2 \nonumber \\
&\leq L \le\|x-x^*_p \ri\|^2_2 + \epsilon  \le\|x-x^*_p \ri\|^2_2 = (L+\epsilon) \le\|x-x^*_p \ri\|^2_2
 \end{align}
 (for $x= x^* \in X^*$ it is trivial and we have an equality), i.e. $g \in \RSI^+(L+\epsilon)$.

Similarly, if $f \in \RSI^-(\mu)$, i.e. $\forall \, x\in\R^d$, $\langle \nabla f(x) , x-x^*_p\rangle  \geq \mu \le\|x-x^*_p \ri\|^2_2  $, consider $g = f+h$ with $h \in \mathcal{F}_{X^*}$, $\|h\|_* = \sup_{x\in\R^d\setminus X^*} \frac{\le\| \nabla h(x) \ri\|_2 }{ d(x,X^*) } < \epsilon <\mu $, then we have $\forall \, x\in \R^d \setminus X^*$
\begin{align}
\langle \nabla g(x) , x-x^*_p\rangle   &= \langle \nabla f(x) + \nabla h(x) , x-x^*_p\rangle  = \langle \nabla f(x) , x-x^*_p\rangle +\langle \nabla h(x) , x-x^*_p \rangle \nonumber \\
&\geq \mu \le\|x-x^*_p \ri\|^2_2 - \le\| \nabla h(x)\ri\|_2 \le\|x-x^*_p \ri\|_2 \nonumber \\
&\geq \mu \le\|x-x^*_p \ri\|^2_2 - \epsilon  \le\|x-x^*_p \ri\|^2_2 = (\mu-\epsilon) \le\|x-x^*_p \ri\|^2_2
 \end{align}
 (for $x= x^*\in X^*$ it is trivial and we have an equality), i.e. $g \in \RSI^-(\mu-\epsilon)$.

 \underline{Continuity of $\EB^-$ and $\EB^+$:}
Given $f \in \EB^+(L)$: $\forall \, x\in\R^d$, $\le\| \nabla f(x) \ri\|_2 \leq L\, d(x,X^*) = L \le\|x-  x^*_p \ri\|_2 $, with $x^*_p \in X^*$ the unique projection of $x$ on $X^*$; this implies
\begin{gather}
\sup_{x\in\R^d\setminus X^*} \dfrac{\le\| \nabla f(x) \ri\|_2 }{ d(x,X^*) } \leq L .
\end{gather}

Given $\epsilon >0$, consider $g \in f + \mathcal{F}_{X^*}$, such that $\|f-g \|_* <\epsilon$: then, 
\begin{align}
\sup_{x\in\R^d\setminus X^*} \dfrac{\le\| \nabla g(x) \ri\|_2 }{ d(x,X^*) }  &\leq \sup_{x\in\R^d\setminus X^*} \dfrac{\le\| \nabla g(x) - \nabla f(x) \ri\|_2 }{ d(x,X^*) }  + \sup_{x\in\R^d\setminus X^* } \dfrac{\le\| \nabla f(x) \ri\|_2 }{ d(x,X^*) }   \leq \epsilon + L
\end{align}
Additionally, since $g \in f+ \mathcal{F}_{x^*}$, $\nabla g(x^*) =0$ $\forall \, x^*\in X^*$, therefore 
\begin{gather}
\le\| \nabla g(x) \ri\|_2 \leq (L+\epsilon) d(x,X^*), \qquad\forall \, x\in\R^d
\end{gather}
i.e. $g \in \EB^+(L+\epsilon)$.

Given $f \in \EB^-(\mu)$: $\forall \, x\in\R^d \ \le\| \nabla f(x) \ri\|_2 \geq \mu \, d(x,X^*) = \mu \|x-x^*_p\|_2 $. 
Fix $\epsilon >0$ and consider $g  \in  f+ \mathcal{F}_{x^*}$, such that $\|f-g \|_*  <\epsilon <\mu$;  in particular $\forall \, x\in \R^d \setminus X^*$, $\|\nabla f(x) - \nabla g(x)\|_2 < \epsilon \, d(x,X^*)$. Then, $\forall \, x\in\R^d\setminus X^*$
\begin{align}
0< (\mu-\epsilon) \, d(x,X^*) &\leq \le\| \nabla f(x) \ri\|_2 - \epsilon d(x,X^*) \nonumber \\
& \leq \le\| \nabla f(x) \ri\|_2- \le\| \nabla f (x) - \nabla g(x)\ri\|_2 \nonumber \\
&\leq \le\| \nabla f(x)  -  \nabla f (x) + \nabla g(x)\ri\|_2 = \le\| \nabla g(x) \ri\|_2
\end{align}
(for $x=x^* \in X^*$ the inequality is trivial), i.e. $g \in \EB^-(\mu-\epsilon)$.

\underline{Continuity of $\PL^-$ and $\PL^+$:}

Let $f\in \PL^-(\mu)$ and $\epsilon >0$. From Figure \ref{assumptions_graph} we have $f\in \QG^-(\mu)$. 
Given $g\in f+\mathcal{F}_{X^*}$ such that $\|f-g\|_*<\mu$, for any $x\in\mathbb{R}^d$ with projection $x_p^*$ onto $X^*$ we have:
\begin{equation}
\|\nabla f(x)-\nabla g(x)\|_2 \leq \|f-g\|_* d\le(x, X^*\ri)
\end{equation}
additionally for $t\in[0,1]$, $d(x_p^*+t(x-x_p^*),X^*)= t\le\|x-x_p^*\ri\|_2$ since $x_p^*\in X^*$ and
\begin{align}
\label{distfg}
|f(x)-g(x)|&=|f(x)-g(x)-(f(x_p^*)-g(x_p^*))| = \le|\int_0^1\langle \nabla(f-g)(x^*_p + t(x-x_p^*)), x-x_p^*\rangle \, \d t \ri|\nonumber\\
&\leq \int_0^1\|f-g\|_*d(x_p^*+t(x-x_p^*), X^*)\|x-x_p^*\|_2 \, \d t\nonumber\\
&\leq \|f-g\|_*\|x-x_p^*\|_2^2\int_0^1 t \, \d t\nonumber\\
&\leq\frac{\|f-g\|_*}{2} \, d\le(x, X^*\ri)^2
\end{align}
Since $f\in\QG^-(\mu)$ and $\|f-g\|_*<\mu$:
\begin{align}
g(x)-g(x_p^*)&\geq f(x)-f^*-|f(x)-g(x)-(f(x_p^*)-g(x_p^*))|\nonumber\\
&\geq \frac{\mu-\|f-g\|_*}{2} \, d(x,X^*)^2\geq 0
\end{align}
Thus $g$ admits a minimum value $g^*$ which is attained at any $x^*\in X^*$. Therefore, $\forall \, x\in \R^d$
\begin{align}
\label{distfg2}
g(x)-g^* \geq \frac{\mu-\|f-g\|_*}{2}\, d(x, X^*)^2.
\end{align}

Since $f\in \PL^-(\mu)$, we have
\begin{align}
\|\nabla g(x)\|_2^2 &=\|\nabla g(x)-\nabla f(x)\|_2^2 + \|\nabla f(x)\|_2 + 2 \langle \nabla g(x)-\nabla f(x) , \nabla f(x) \rangle \nonumber \\
&\geq 0 + 2\mu(f(x)-f^*) - 2\|\nabla g(x)-\nabla f(x)\|_2 \sqrt{2\mu(f(x)-f^*)} \nonumber \\
&\geq 2\mu(g(x)-g^*) - 2\mu|f(x)-g(x)-(f^*-g^*)| \nonumber \\
&\quad - 2\|f-g\|_* \, d(x,X^*)\sqrt{2\mu(g(x)-g^*+|f(x)-g(x)-(f^*-g^*)|)}
\end{align}

The second term can be easily bounded by \eqref{distfg} and \ref{distfg2}; the third term can be bounded as follows
\begin{align}
    \sqrt{g(x)-g^*+|f(x)-g(x)-(f^*-g^*)|} &\leq \sqrt{(g(x)-g^*)}+\sqrt{|f(x)-g(x)-(f^*-g^*)|}\nonumber\\
    &\leq \sqrt{(g(x)-g^*)} + \sqrt{\frac{\|f-g\|_*}{2}d(x,X^*)^2} \nonumber \\
     &\leq \sqrt{(g(x)-g^*)} + \sqrt{\frac{\|f-g\|_*}{\mu-\|f-g\|_*}(g(x)-g^*)} \nonumber \\
&= \le( 1 + \sqrt{\frac{\|f-g\|_*}{\mu-\|f-g\|_*}}\ri) \sqrt{(g(x)-g^*)}
\end{align}
where we applied again \eqref{distfg} and \eqref{distfg2}.
Finally we get:
\begin{align}
    \le\|\nabla g(x)\ri\|_2^2&\geq 2\le[\mu-\frac{\mu\|f-g\|_*}{\mu-\|f-g\|_*}-\|f-g\|_*\sqrt{\frac{2}{\mu-\|f-g\|_*}}\le(1+\sqrt{\frac{\|f-g\|_*}{\mu-\|f-g\|_*}}\ri)\ri](g(x)-g^*) \nonumber \\
    &\geq 2(\mu-\epsilon) (g(x)-g^*)
\end{align}
provided that $\|f-g\|_*$ is small enough. Indeed, the quantity 
\begin{gather*}
 0\leq  \frac{\mu\|f-g\|_*}{\mu-\|f-g\|_*}+ \|f-g\|_*\sqrt{\frac{2}{\mu-\|f-g\|_*}}\le(1+\sqrt{\frac{\|f-g\|_*}{\mu-\|f-g\|_*}}\ri) \to 0, \quad \text{as } \|f-g\|_* \to 0,
\end{gather*}
therefore $\forall \, \epsilon >0$, $\exists \, \delta >0$ such that for $\|f-g\|_* \leq \delta$, we have
\begin{gather*}
  \frac{\mu\|f-g\|_*}{\mu-\|f-g\|_*}+ \|f-g\|_*\sqrt{\frac{2}{\mu-\|f-g\|_*}}\le(1+\sqrt{\frac{\|f-g\|_*}{\mu-\|f-g\|_*}}\ri) \leq \epsilon.
\end{gather*}
In conclusion, $g\in \PL^-(\mu-\epsilon)$.


Let us now consider $f\in \PL^+(L)\cap \QG^-(\mu)$, and $g\in f+\mathcal{F}_{X^*}$ such that $||f-g||_*<\mu$. 
\begin{gather}
    \le\|\nabla g(x)\ri\|_2^2 = \le\|\nabla g(x)-\nabla f(x)\ri\|_2^2 + \le\|\nabla f(x)\ri\|_2^2 + 2 \langle \nabla g(x)-\nabla f(x),  \nabla f(x) \rangle
\end{gather}
The second term can be estimated thanks to \eqref{distfg} and \eqref{distfg2}:
\begin{align}
    \le\|\nabla f(x)\ri\|_2^2 &\leq 2L(f(x)-g(x)-(f^*-g^*))+2L(g(x)-g^*)\nonumber\\
    &\leq 2L\le(\frac{||f-g||_*}{\mu-||f-g||_*}+1\ri)(g(x)-g^*);
\end{align}
and similarly the third term:
\begin{align}
    \langle \nabla g(x)-\nabla f(x), \nabla f(x)\rangle &\leq \le\|\nabla g(x)-\nabla f(x)\ri\|_2\le\|\nabla f(x)\ri\|_2\nonumber\\
    &\leq ||f-g||_*d(x,X^*)\sqrt{2L\le(\frac{||f-g||_*}{\mu-||f-g||_*}+1\ri)(g(x)-g^*)}\nonumber\\
    &\leq ||f-g||_*\sqrt{\frac{4L}{\mu-||f-g||_*}\le(\frac{||f-g||_*}{\mu-||f-g||_*}+1\ri)}(g(x)-g^*).
\end{align}
This finally leads to:
\begin{align}
    \le\|\nabla g(x)\ri\|_2^2 
    &\leq 2(L + K)(g(x)-g^*) \leq 2(L+\epsilon) (g(x)-g^*)
\end{align}
where 
$$K=\le[\frac{\|f-g\|_*}{\mu-\|f-g\|_*}+\frac{L\|f-g\|_*}{\mu-\|f-g\|_*}+\|f-g\|_*\sqrt{\frac{4L}{\mu-\|f-g\|_*}\le(\frac{\|f-g\|_*}{\mu-\|f-g\|_*}+1\ri)}\ri],$$
provided that $\|f-g\|_*$ is small enough.
Following a similar argument as before, we can easily see that $K\geq 0$ and $K \to 0$ as $\|f-g\|_*\to 0$, therefore $\forall \, \epsilon >0$, $\exists \, \delta >0$ such that if $\|f-g\|_* \leq\delta$, then $K\leq \epsilon$. Therefore, $g \in \PL^+(L+\epsilon)$.

\underline{Continuity of $\QG^-$ and $\QG^+$:}

Given $f \in \QG^+(L)$: $f(x) - f^* \leq  \frac{L}{2} \, d(x,X^*)^2$, $\forall \, x\in\R^d$. Consider $g = f+h$, $h \in\mathcal{F}_{X^*}$ with $ \|h\|_* = \sup \frac{\|\nabla f(x) - \nabla g(x)\|_2}{d(x,X^*)} =  \sup \frac{\|\nabla h(x) \|_2}{d(x,X^*)} \leq \epsilon$, then $\forall \, x \in \R^d$, with $x^*_p \in X^*$ the corresponding projection on $X^*$,
\begin{align}
g(x) - g^* &= f(x) + h(x) - (f^*+h^*) = f(x) - f^* + h(x) - h^* \nonumber\\
&\leq \frac{L}{2}d(x,X^*)^2 + \int_0^1\langle \nabla h(x^*_p + t(x-x^*_p) ), x-x^*_p\rangle \, \d t \nonumber \\
&\leq \frac{L}{2}d(x,X^*)^2 + \int_0^1 \le\| \nabla h(x^*_p + t(x-x^*_p) )\ri\|_2 \| x-x^*_p\|_2 \, \d t \nonumber \\
&\leq \frac{L}{2}d(x,X^*)^2 + \int_0^1 \dfrac{\le\| \nabla h(x^*_p + t(x-x^*_p) )\ri\|_2}{\|x^*_p + t(x-x^*_p) - X^* \|_2} \|x^*_p + t(x-x^*_p) - X^* \|_2 \| x-x^*_p\|_2 \, \d t \nonumber \\
&\leq  \frac{L}{2}d(x,X^*)^2  + \|h\|_* \|x-x^*_p\|_2^2 \int_0^1 t \, \d t \nonumber \\
&\leq  \frac{L+\epsilon}{2}d(x,X^*)^2  
\end{align}
where we used $ \|x^*_p + t(x-x^*_p) - X^* \|_2 \leq t  \|x-x^*_p \|_2$, $\forall\, t\in[0,1] $; as before, for $x = x^* \in X^*$ the inequality is trivial. Therefore, $g \in \QG^+(L+\epsilon)$.

The proof that $g \in \QG^-(\mu-\epsilon)$ if $f\in \QG^-(\mu)$ for $g\in f + \mathcal{F}_{X^*}$, $\|f-g\|_* \leq\epsilon <\mu$, follows the same argument.

    \section{Graph of lower conditions}\label{App_graph_lower}
    \paragraph{\underline{$\SC^-(\mu) \rightarrow ~^*\!\SC^-(\mu)$:}}

Immediate by taking $y= x^*_p$ (the projection of $x\in \R^d$ onto $X^*$) in the definition of strong convexity.

\paragraph{\underline{$^*\!\SC^-(\mu) \rightarrow \PL^-(\mu)$:}}

Assume $f \in ~^*\!\SC^-(\mu)$:  $f^* \geq 
    f \left( x \right)
    + \langle \nabla f ( x ) , x^*_p - x\rangle
    + \frac{\mu}{2} \left\| x^*_p - x \right\|_2^2$, $\forall \, x \in \R^d$. Hence,
\begin{gather}
    f^* - f \left( x \right) \geq
    - \frac{1}{2 \mu} \left\| \nabla f(x) \right\|_2^2
    + \frac{1}{2 \mu} \left\| \nabla f(x) + \mu \left( x^*_p - x \right) \right\|^2_2
    \geq 
    - \frac{1}{2 \mu} \left\| \nabla f(x) \right\|^2_2
\end{gather}
i.e. $\left\| \nabla f(x) \right\|^2_2 \geq  2 \mu ( f - f^* )$. Therefore, $f \in \PL^-(\mu)$.

\paragraph{\underline{$\PL^-(\mu) \rightarrow \QG^-(\mu)$:}}

The claim was originally proven in \citep{DBLP:journals/corr/KarimiNS16}, following some arguments from \citep{bolte2017error} and \citep{zhang2017restricted} and we will report it here for the sake of completeness.

Consider the gradient flow of $g(x) = \sqrt{f(x) - f^*}$: $x'(t) = - \nabla g(x(t))$. 
Note the $f \in \PL^-(\mu)$ implies that $\| \nabla g(x) \|_2^2 \geq \frac{\mu}{2}>0$ $\forall \, x\in\R^d$; in particular, despite the fact that $g$ attains its minimum on the set $X^*$, $\nabla g$ may not be defined on $X^*$ and the gradient flow equation ceases to be defined once $X^*$ is reached.
We then study the path of a gradient flow of $g$ until it hits $X^*$:
$\forall \, x_0 \in \R^d$, $\forall \, T>0$ for which the flow is defined,
\begin{align}
    g(x_0) \geq &~ g(x_0) - g(x_T) 
    =  - \int_{0}^{T} \langle \nabla g(x(t)) ,  x'(t)\rangle   \, \d t 
    =  \int_{0}^{T} \left\| \nabla g(x(t)) \right\|^2_2 \, \d t \nonumber \\
    \geq &~ \int_{0}^{T} \frac{\mu}{2} \, \d t = \frac{\mu}{2} T ,
\end{align}
where the first inequality follows from the fact that $g$ is non-negative and the second inequality follows from the $\PL^-(\mu)$ property.
 This proves the existence of $  T^* =  T^*(x_0)$ such that $x_{T^*} \in X^*$. 
 
Therefore, $\forall \, x_0 \in \R^d$
\begin{align}
    g(x_0) = & g(x_0) - g(x_{T^*})= \int_{0}^{T^*} \left\| \nabla g(x(t)) \right\|^2_2 \, \d t\nonumber\\
    \geq & \sqrt{\frac{\mu}{2}} \int_{0}^{T^*} \left\| \nabla g(x(t)) \right\|_2 \, \d t 
    =  \sqrt{\frac{\mu}{2}} \int_{0}^{T^*} \left\|  x'(t) \right\|_2 \, \d t \nonumber\\ 
    \geq & \sqrt{\frac{\mu}{2}} \le\| \int_{0}^{T^*}  x'(t)  \, \d t \ri\|_2 = \sqrt{\frac{\mu}{2}} \left\| x_0 - x_{T^*} \right\|_2 \nonumber \\
    \geq & \sqrt{\frac{\mu}{2}} d(x_0, X^*) 
\end{align}
and, by squaring on both sides,
\begin{equation}
    f(x) - f^* = g(x)^2 \geq \frac{\mu}{2} d(x,X^*)^2;
\end{equation}
i.e. $f \in \QG^-(\mu)$.

\paragraph{\underline{$^*\!\SC^-(\mu_1) \mathrm{~ and ~} \QG^-(\mu_2) \rightarrow \RSI^-\left(\frac{\mu_1 + \mu_2}{2}\right)$:}}

For $f \in ~^*\!\SC^-(\mu_1) \cap \QG^-(\mu_2)$, we have
\begin{gather}
    \langle \nabla f ( x ) , x - x^*_p \rangle
    \geq 
    f(x) - f^*
    + \frac{\mu_1}{2} \left\| x^*_p - x \right\|^2_2
    \geq 
    \frac{\mu_1 +\mu_2}{2} \left\| x^*_p - x \right\|^2_2
\end{gather}
i.e. $f \in \RSI^-\left(\frac{\mu_1 + \mu_2}{2}\right)$.
Note that this holds also for non positive $\mu_1$.
In particular, if $f \in \QG^-(\mu)$ and $f$ is *-convex ($\mu_1=0$), then $f \in \RSI^-(\frac{\mu}{2})$.

\paragraph{\underline{$^*\!\SC^-(\mu) \rightarrow \RSI^-\left( \mu \right)$:}}

This follows directly from the three previous results.
Indeed, $^*\!\SC^-(\mu) \subseteq \PL^-(\mu) \subseteq \QG^-(\mu)$ and $^*\!\SC^-(\mu) \cap \QG^-(\mu)\subseteq\RSI^-( \mu )$.

\paragraph{\underline{$\RSI^-(\mu) \rightarrow \QG^-(\mu)$:}}
For every $x\in \R^d$ consider the line segment $x(t) = x^*_p + t(x - x^*_p)$, $t\in [0,1]$,  with $x^*_p\in X^*$ the projection of $x$ onto $X^*$. It is clear that $\forall \, t\in[0,1]$ the projection of $x(t)$ onto $X^*$ is still $x^*_p$. Since $f \in \RSI^-(\mu)$, $\forall \, x\in \R^d$
\begin{gather}
    \langle \nabla f(x^*_p + t(x - x^*_p)), t(x-x^*_p)\rangle \geq \mu \|t (x-x^*_p)\|_2^2 = \mu t^2 \| (x-x^*_p)\|_2^2,
\end{gather}
therefore
\begin{align}
    f(x) - f^* = \int_0^1 \langle \nabla f(x^*_p + t(x-x^*_p)) , x-x^*_p\rangle \, \d t  \geq  \int_0^1 \mu t \left\| x - x^*_p \right\|^2_2 \, \d t = \frac{\mu}{2} \left\| x - x^*_p \right\|^2_2,
\end{align}
implying that $f\in \QG^-(\mu)$.

\paragraph{\underline{$\RSI^-(\mu) \rightarrow \EB^-(\mu)$:}}
It follows from Cauchy-Schwartz inequality.

\paragraph{\underline{$\PL^-(\mu_1) \cap \QG^-(\mu_2) \rightarrow \EB^-\left(\sqrt{\mu_1 \mu_2}\right)$:}}
Assume $f \in \PL^-(\mu_1) \cap \QG^-(\mu_2)$:
\begin{gather}
    \frac{1}{2} \left\| \nabla f(x) \right\|^2_2 \geq  \mu_1 \left( f(x) - f^* \right)  \geq \frac{\mu_1\mu_2}{2} \| x - x^*_p \|^2 _2
\end{gather}
i.e. $\| \nabla f(x) \|_2 \geq  \sqrt{\mu_1\mu_2} \| x - x^*_p \|_2$.

Hence, $f \in \EB^-\left(\sqrt{\mu_1 \mu_2}\right)$.
Note that $\PL^-(\mu) \subseteq \QG^-(\mu)$, therefore $\PL^-(\mu) \subseteq \EB^-(\mu)$ (set $\mu_1=\mu_2=\mu$).

\paragraph{\underline{$\EB^-(\mu) \cap \QG^+(L) \rightarrow \PL^-(\mu^2/L)$:}}
Given $f\in \EB^-(\mu) \cap \QG^+(L)$, $\forall \, x\in\R^d$
\begin{gather}
    \| \nabla f(x) \|^2_2 \geq  \mu^2 \| x - x^*_p \|_2^2
    \geq  \frac{2\mu^2}{L} \left( f(x) - f^* \right)
\end{gather}
i.e. $f\in \PL^-(\mu^2/L)$.

    \section{Graph of upper conditions}\label{App_graph_upper}

\paragraph{\underline{$\SC^+(L) \rightarrow \PL^+(L)$:}}

Assume $f \in \SC^+(L)$, hence $\forall \, x,y\in R^d$
\begin{align}
    f(y) &\leq 
    f \left( x \right)
    +\langle  \nabla f ( x ) , y - x\rangle
    + \frac{L}{2} \left\| y - x \right\|^2_2  \nonumber\\
   & = 
    f(x)
    - \frac{1}{2 L} \left\| \nabla f(x) \right\|^2_2
    + \frac{1}{2 L} \left\| \nabla f(x) + L \left( y - x \right) \right\|^2_2
\end{align}
In particular, $\forall \, x,y \in \R^d$
\begin{align}
    f^* \leq f(y) \leq 
    f(x)
    - \frac{1}{2 L} \left\| \nabla f(x) \right\|^2_2
    + \frac{1}{2 L} \left\| \nabla f(x) + L \left( y - x \right) \right\|^2_2
    \end{align}
    and by choosing $y = x - \frac{\nabla f(x)}{L}$, we have
\begin{gather}
    f^* - f(x)
    \leq 
    - \frac{1}{2 L} \left\| \nabla f(x) \right\|^2_2, \quad \text{i.e.} \quad 
   \frac{1}{2} \left\| \nabla f(x) \right\|^2_2 \leq   L \left( f(x) - f^* \right)
\end{gather}
Hence, $f \in \PL^+(L)$.

\paragraph{\underline{$\PL^+(L) \rightarrow ~^*\!\SC^+(L)$:}}

Assume $f \in \PL^+(L)$, hence
\begin{align}
    f^* - f(x)
    \leq &~
    - \frac{1}{2 L} \left\| \nabla f(x) \right\|^2_2 \nonumber \nonumber \\
    \leq &~
    - \frac{1}{2 L} \left\| \nabla f(x) \right\|^2_2
    + \frac{1}{2 L} \left\| \nabla f(x) + L \left( x^*_p - x \right) \right\|^2_2 \nonumber \\
    f^* \leq &~
    f \left( x \right)
    + \langle \nabla f ( x ) , x^*_p - x\rangle
    + \frac{L}{2} \left\| x^*_p - x \right\|^2_2
\end{align}
Hence, $f \in ~^*\!\SC^+(L)$.

\paragraph{\underline{$\PL^+(L) \rightarrow \QG^+(L)$:}}

Assume $f \in \PL^+(L)$ and consider the function $g(x) = \sqrt{f(x) - f^*}$: since $f \in \PL^+(L)$, we have $\|\nabla g(x)\|_2^2 \leq \frac{L}{2} $, $\forall \, x \in\R^d$. Then,
\begin{align}
    g(x) = & g(x) - g(x^*_p) 
    =  \int_0^1\langle  \nabla g(x^*_p + t(x-x^*_p)) ,x-x^*_p\rangle  \, \d t \nonumber \\
    \leq & \int_0^1  \le\|\nabla g(x^*_p + t(x-x^*_p))\ri\|_2 \le\|( x-x^*_p)\ri\|_2  \, \d t \nonumber \\ 
    \leq & \int_0^1 \sqrt{\frac{L}{2}} \left\| x-x^*_p \right\|_2 \, \d t 
    \leq  \sqrt{\frac{L}{2}} \left\| x-x^*_p \right\|_2
\end{align}
Therefore, by squaring on both sides, 
\begin{gather}
    f(x) - f^* \leq \dfrac{L}{2} \|x-x^*_p \|_2^2.
\end{gather}

Note: this result is not explicit in the graph as it can be recover by following the existing edges. However we needed to prove it here for the following result. 

\paragraph{\underline{$\PL^+(L) \rightarrow \EB^+(L)$:}}
Assume $f \in \PL^+(L_1) \cap \QG^+(L_2)$, then 
\begin{gather}
    \| \nabla f(x) \|^2_2
    \leq 2 L_1 (f(x) - f^*) \leq L_1L_2 \| x-x^*_p \|^2_2,
\end{gather}
hence  $f \in \EB^+(\sqrt{L_1 L_2})$.
In particular, from the previous result we have that if $f \in \PL^+(L)$, then $f \in \QG^+(L)$, hence $f \in \EB^+(L)$ (take $L_1=L_2=L$).

\paragraph{\underline{$\EB^+(L) \rightarrow \RSI^+(L)$:}}

Given $f \in \EB^+(L)$,
\begin{gather}
     \langle \nabla f(x) , x - x^*_p\rangle
    \leq  \| \nabla f(x) \|_2 \cdot \| x - x^*_p \|_2
    \leq  L \| x - x^*_p \|^2_2,
\end{gather}
therefore $f \in \RSI^+(L)$.

\paragraph{\underline{$^*\!\SC^+(L) \rightarrow \QG^+(L)$:}}
For each $x \in \R^d$, with $x^*_p\in X^*$ its projection onto $X^*$,  define
$$g(t) = \frac{\frac{L}{2} \| t (x - x^*_p) \|_2^2 - \left( f(x^*_p + t(x-x^*_p)) - f^* \right)}{t}, \qquad t\in (0,+\infty).$$
We verify that 
\begin{gather}
    g'(t)
    = 
    \frac{\frac{L}{2} \left\| t (x - x^*_p) \right\|^2_2 - \langle \nabla f(x^*_p + t(x-x^*_p)) ,  x - x^*_p\rangle + \left( f(x^* + t(x-x^*_p)) - f^* \right)}{t^2} 
    \geq  0
\end{gather}
since $f \in ~^*\!\SC^+(L)$. Therefore, $g$ is monotonically increasing on $(0,+\infty)$.
Additionally, $g$ can be continuously extended in
$t=0$ by l'H\^{o}pital's rule:
$$ \lim_{t\to 0_+} g(t) = \lim_{t\to 0_+} Lt \| (x - x^*_p) \|_2^2 - \langle \nabla f(x^*_p + t(x-x^*_p)), x-x^*_p\rangle  = 0. $$

Therefore,
$$g(1) = \frac{L}{2} \|  x - x^*_p \|_2^2 - \left( f(x) - f^* \right) \geq g(0) = 0$$
i.e. $ f(x) - f^*   \leq \frac{L}{2} \|  x - x^*_p \|_2^2  $: $f \in \QG^+(L)$.

\paragraph{\underline{$^*\!\SC^+(L) \rightarrow \RSI^+(L)$:}}
Let $f \in ~^*\!\SC^+(L_1) \cap \QG^+(L_2)$: 
\begin{align}
    \langle \nabla f(x) ,  x - x^*_p\rangle
    \leq &
    f(x) - f^* + \frac{L_1}{2} \left\| x - x^*_p \right\|^2_2 \leq \frac{L_1+L_2}{2} \left\| x - x^*_p \right\|^2_2,
\end{align}
therefore $f \in \RSI^+(\frac{L_1+L_2}{2})$.
In particular, since $^*\!\SC^+(L) \subseteq \QG^+(L)$, then $^*\!\SC^+(L) \subseteq \RSI^+(L)$.

\paragraph{\underline{$\RSI^+(L) \rightarrow ~^*\!\SC^+(2L)$:}}
For $f \in \RSI^+(L)$, we have
\begin{gather}
    \langle \nabla f(x) ,  x - x^*_p \rangle
    \leq 
    L \left\| x - x^*_p \right\|^2_2 
    \leq 
    f(x) - f^* + L \left\| x - x^*_p \right\|^2_2
\end{gather}
i.e. $f \in ~^*\!\SC^+(2L)$.

\paragraph{\underline{$\RSI^+(L) \rightarrow \QG^+(L)$:}}
For every $x\in \R^d$ consider the line segment $x(t) = x^*_p + t(x - x^*_p)$, $t\in [0,1]$; recall that $\forall \, t\in[0,1]$ the projection of $x(t)$ onto $X^*$ is still $x^*_p$. Since $f \in \RSI^+(L)$, $\forall \, x\in \R^d$
\begin{gather}
    \langle \nabla f(x^*_p + t(x - x^*_p)), t(x-x^*_p) \rangle \leq L \|t (x-x^*_p)\|_2^2 = L t^2 \| x-x^*_p\|_2^2,
\end{gather}


Therefore,  $f \in \QG^+(L)$:
\begin{gather}
    f(x) - f^* =  \int_0^1 \langle \nabla f(x^* + t(x - x^*_p)) , x-x^*_p\rangle \, \d t \leq  \int_0^1 L t \left\| x - x^*_p \right\|^2_2 \ \d t = \frac{L}{2} \left\| x - x^*_p \right\|^2_2.
\end{gather}

\paragraph{\underline{$\SC^-(\mu) \mathrm{~and~} \QG^+(L) \rightarrow \EB^+\left(L + \sqrt{L(L - \mu)}\right)$:}}

Assume $f \in \SC^-(\mu) \cap \QG^+(L)$, with $\mu<L$, and $\mu$ can be non positive (we recall that $f \in \SC^-(0)$ is convex). The case $\mu \geq L$ is trivial as it implies $f(x) - f^* = \frac{L}{2}\|x - x^*_p\|^2$  $\, \forall \, x\in\R^d$. 

We have by definition: $\forall \, x,y \in \R^d$
\begin{gather}
    f(x) - f^* +\langle  \nabla f(x) ,  y - x\rangle  + \frac{\mu}{2}\left\| y - x \right\|_2^2 \stackrel{\SC^-}{\leq} f(y) - f^* \stackrel{\QG^+}{\leq}  \frac{L}{2} \left\| y - y^*_p \right\|^2_2 \leq  \frac{L}{2} \left\| y - x^*_p \right\|^2_2;
\end{gather}
in particular, 
\begin{gather}
    f(x) - f^* + \langle \nabla f(x) ,  y - x\rangle + \frac{\mu}{2}\left\| y - x \right\|_2^2  \leq  \frac{L}{2} \left\| y - x^*_p \right\|^2_2
\end{gather}
and by choosing $y = \frac{Lx^*_p - \mu x + \nabla f(x)}{L-\mu}$ we have
\begin{align}
    L \mu \left\| x - x^*_p \right\|^2_2 + \left\| \nabla f(x) \right\|^2_2 + 2L \langle \nabla f(x) ,  x^*_p - x\rangle
    \leq &~
    2(L - \mu) \cdot (f^* - f(x)) \label{eq:SC-QG+}
\end{align}

The RHS is non positive, then by removing it and factoring the LHS
\begin{gather}
    \left\| \nabla f(x) + L(x^*_p - x) \right\|^2_2
    \leq 
    L(L - \mu) \left\| x - x^*_p \right\|^2_2;
    \end{gather}
finally, by triangle inequality,
\begin{align}
    \left\| \nabla f(x) \right\|_2 - L \left\| x^*_p - x \right\|_2
    \leq &~
    \sqrt{L(L - \mu)} \left\| x - x^*_p \right\|_2\\ 
    \left\| \nabla f(x) \right\|_2
    \leq &~
    \left( L + \sqrt{L(L - \mu)} \right) \left\| x - x^*_p \right\|_2
\end{align}

Hence, $f \in \EB^+\left(L + \sqrt{L(L - \mu)}\right)$.
Note that for $\mu = 0$ (i.e. $f$ is convex), we have $\QG^+(L) \rightarrow \EB^+(2L)$.

\paragraph{\underline{$\SC^-(\mu) \mathrm{~and~} ~^*\!\SC^+(L) \rightarrow \EB^+\left( L + 2\max\{-\mu, 0\} \right)$:}}

Assume $f \in \SC^-(\mu) \cap ~^*\!\SC^+(L)$.
In particular $f \in \QG^+(L)$, then all the previous results still hold. From \eqref{eq:SC-QG+} we have 
\begin{gather}
    L \mu \left\| x - x^*_p \right\|^2_2 + \left\| \nabla f(x) \right\|^2_2 + 2L\langle  \nabla f(x) ,  x^*_p - x \rangle
    \leq 
    2(L - \mu) \cdot (f^* - f(x))\nonumber \\
    \leq 
    2(L - \mu) \cdot \left[ \langle \nabla f(x) ,  x^*_p - x\rangle  + \frac{L}{2}\left\| x - x^*_2 \right\|^2_2 \right] 
\end{gather}
thanks to $f \in ~^*\!\SC^+(L)$, i.e.
$$\left\| \nabla f(x) \right\|^2_2 + 2\mu\langle  \nabla f(x) ,  x^*_p - x\rangle \leq
     L (L - 2\mu) \left\| x - x^*_p \right\|^2_2.$$
After rearranging the terms, we obtain
$\left\| \nabla f(x) + \mu(x^*_p - x) \right\|^2_2
    \leq 
    (L - \mu)^2 \left\| x - x^*_p \right\|^2_2 $
    and by triangle inequality
    \begin{gather}
    \left\| \nabla f(x) \right\|_2 - |\mu| \left\| x^*_p - x \right\|_2
    \leq 
    (L - \mu) \left\| x - x^*_p \right\|_2,
\end{gather}
i.e.
$$ \left\| \nabla f(x) \right\|_2
    \leq 
    (L + 2\max\{-\mu, 0\}) \left\| x - x^*_p \right\|_2. $$
Finally $f \in \EB^+\left( L + 2\max\{-\mu, 0\} \right)$.
In particular, under convex assumption ($\mu = 0$),
$^*\!\SC^+(L) \rightarrow \EB^+(L)$.

\paragraph{\underline{$\QG^-(\mu) \mathrm{~and~} \EB^+(L) \rightarrow \PL^+\left( \frac{L^2}{\mu} \right)$:}}

Let $f \in \QG^-(\mu) \cap \EB^+(L)$, we have:
\begin{gather}
    \frac{1}{2} \left\| \nabla f(x) \right\|^2_2
    \leq 
    \frac{1}{2} L^2 \left\| x - x^*_p \right\|^2_2 
    \leq 
    \frac{1}{2} L^2 \frac{2}{\mu} (f(x) - f^*)=
    \frac{L^2}{\mu} (f(x) - f^*),
\end{gather}
therefore, $f \in \PL^+\left( \frac{L^2}{\mu} \right)$.

    \section{Rates of convergence}\label{App_rates_conv}
    \paragraph{\underline{Under $\SC^-(\mu)$ and $\SC^+(L)$}}

This is a known result and we refer to the proof in \citep[Section 3.4.2]{bubeck2014convex}.
Let's assume $f \in \SC^-(\mu) \cap \SC^+(L)$ with $L>\mu$ (the other case is trivial): $\forall \, x, y, z \in \R^d$

\begin{equation}
    f(y) + \langle \nabla f(y) ,  x - y\rangle + \frac{\mu}{2} \left\| x - y \right\|^2_2
    \overset{\SC^-(\mu)}{\leq} f(x) \overset{\SC^+(L)}{\leq}
    f(z) + \langle \nabla f(z) ,  x - z\rangle + \frac{L}{2} \left\| x - z \right\|^2_2
    \label{eq:stronvandsmooth}
\end{equation}
i.e. $\forall \, x,y,z \in\R^d$
\begin{equation}
    f(z)  - f(y) +\langle  \nabla f(z) , x - z\rangle - \langle \nabla f(y) ,  x - y\rangle + \frac{L}{2} \left\| x - z \right\|^2_2   - \frac{\mu}{2} \left\| x - y \right\|^2_2 \geq 0.
\end{equation}
By minimizing the left hand side of the above expression with respect to the variable $x$, we find that for 
\begin{gather}
    x = \frac{L z - \mu y + \nabla f(y) - \nabla f(z)}{L - \mu}
\end{gather}
the inequality becomes


\begin{equation}
     f(y) - f(z) \leq \frac{1}{L - \mu} \left[ \langle z - y,   \mu \nabla f(z) - L \nabla f(y)\rangle - \frac{1}{2}\left\| \nabla f(y) - \nabla f(z) \right\|_2^2 - \frac{L\mu}{2} \left\| y - z \right\|_2^2 \right]
\end{equation}
 $\forall \, y, z \in \R^d$.
%
By swapping the roles of $y$ and $z$, summing, and rearranging terms, we obtain the well-known inequality (see, e.g.\ ~\citep{Nesterov2}):  $\forall \, y, z \in \R^d$
\begin{equation}
     \langle z - y,  \nabla f(z) - \nabla f(y)\rangle  \geq \frac{1}{L + \mu} \left( \left\| \nabla f(y) - \nabla f(z) \right\|^2 + L\mu \left\| y - z \right\|^2 \right). \label{eq:generalized_cocoercivity}
\end{equation}

Note that $f \in \SC^-(\mu)$ implies that $X^*=\left\{x^*\right\}$. In conclusion,
\begin{align}
    \left\| x_{n+1} - x^* \right\|^2_2 =
    &~ \left\| x_n - x^* -\alpha \nabla f(x_n) \right\|^2_2 \nonumber \\
    &~ \left\| x_n - x^* \right\|^2_2 - 2 \alpha  \langle \nabla f(x_n) ,  x_n - x^*\rangle  + \alpha^2 \left\| \nabla f(x_n) \right\|^2_2 \nonumber \\
    \leq &~ \left( 1 - \frac{2\alpha L\mu}{L + \mu} \right) \left\| x_n - x^* \right\|^2_2 + \alpha \left( \alpha - \frac{2}{L + \mu} \right) \left\| \nabla f(x_n) \right\|^2_2 \nonumber \\
    = &~ \left( \frac{\kappa - 1}{\kappa + 1} \right)^2 \left\| x_n - x^* \right\|^2_2
\end{align}
for $\alpha = \frac{2}{L + \mu}$.

\paragraph{\underline{Under $\PL^-(\mu)$ and $\SC^+(L)$}} 

Let's assume $f \in \PL^-(\mu) \cap \SC^+(L)$.
Then,
\begin{align}
    f(x_{n+1}) - f^*
    \leq &~ f(x_n) - f^* - \alpha \left( 1 - \frac{L\alpha}{2} \right) \left\| \nabla f(x_n) \right\|^2_2 \nonumber\\
    \leq &~ f(x_n) - f^* -2\mu \alpha \left( 1 - \frac{L\alpha}{2} \right)  \left( f(x_n) - f^* \right) \nonumber \\
    = &~ \left( 1 - \frac{1}{\kappa} \right) \left( f(x_n) - f^* \right)
\end{align}
for $\alpha = \frac{1}{L}$.

\paragraph{\underline{Under $~^*\!\SC^-(\mu)$ and $\PL^+(L)$}}

Assume $f \in ~^*\!\SC^-(\mu) \cap \PL^+(L)$.  $\forall \ n\in\mathbb{N}$, let $x_{n,p}^*$ be the projection of $x_n$ on $X^*$.
Then,
\begin{align}
    d(x_{n+1},X^*)^2\leq\left\| x_{n+1} - x^*_{n,p} \right\|^2_2
    = &~ \left\| x_n - x^*_{n,p} \right\|^2_2 - 2 \alpha \langle x_n - x^*_{n,p},  \nabla f(x_n) \rangle + \alpha^2 \left\| \nabla f(x_n) \right\|^2_2 \nonumber \\
    \leq &~ \left\| x_n - x^*_{n,p} \right\|^2 - 2 \alpha \left( f(x_n) - f^* + \frac{\mu}{2} \left\| x_n - x^*_{n,p} \right\|^2_2 \right) \nonumber \\
    &~ \hspace{1.8cm} + 2\alpha^2 L \left( f(x_n) - f^* \right) \nonumber \\
    = &~ \left( 1 - \mu\alpha \right) \left\| x_n - x^*_{n,p} \right\|^2_2 - 2\alpha (1 - L\alpha) (f(x_n) - f^*) \\
    = &~ \left( 1 - \frac{1}{\kappa} \right) d(x_n, X^*)^2 \tag*{for $\alpha = \frac{1}{L}$.}
\end{align}

Note this proof is quite similar to the proof of Theorem 3.1 in \citep{gower2019sgd} applied directly to the deterministic case.

Next, we show a similar proof that follows the same idea but doesn't require $~^*\!\SC^-(\mu)$.

\paragraph{\underline{Under $~^*\!\SC^-(0)$, $\RSI^-(\mu)$ and $\PL^+(L)$}}

Assume $f \in ~^*\!\SC^-(0) \cap \RSI^-(\mu) \cap \PL^+(L)$.
From star convexity, we have $\langle \nabla f(x) , x - x_p \rangle \geq f(x) - f^*$, and from restricted secant inequality $\langle \nabla f(x) , x - x_p \rangle \geq \mu \|x - x_p \|^2$.
Combining the two, we obtain $$\langle \nabla f(x) , x - x_p \rangle \geq \frac{1}{2}\left( f(x) - f^* \right) + \frac{\mu}{2} \|x - x_p \|^2.$$

With similar argument as above, denote $x_{n,p}^*$ the projection of $x_n$ on $X^*$, $\forall \, n \in \mathbb{N}$.
Then,
\begin{align}
    d(x_{n+1},X^*)^2\leq\left\| x_{n+1} - x^*_{n,p} \right\|^2_2
    = &~ \left\| x_n - x^*_{n,p} \right\|^2_2 - 2 \alpha \langle x_n - x^*_{n,p},  \nabla f(x_n) \rangle + \alpha^2 \left\| \nabla f(x_n) \right\|^2_2 \nonumber \\
    \leq &~ \left\| x_n - x^*_{n,p} \right\|^2 - 2 \alpha \left( \frac{1}{2}\left( f(x_n) - f^* \right) + \frac{\mu}{2} \left\| x_n - x^*_{n,p} \right\|^2_2 \right) \nonumber \\
    &~ \hspace{1.8cm} +  2\alpha^2 L \left( f(x_n) - f^* \right) \nonumber \\
    = &~ \left( 1 - \mu\alpha \right) \left\| x_n - x^*_{n,p} \right\|^2_2 - \alpha (1 - 2L\alpha) (f(x_n) - f^*) ]\nonumber \\
    = &~ \left( 1 - \frac{1}{2\kappa} \right) d(x_n, X^*)^2 
\end{align}
for $\alpha = \frac{1}{2L}$.

\paragraph{\underline{Under $\RSI^-(\mu)$ and $\EB^+(L)$}}

Assume $f \in \RSI^-(\mu) \cap \EB^+(L)$, and for some $n\in\mathbb{N}$, $x_p^*$ denotes the projection of $x_n$ on $X^*$.
Then
\begin{align}
    d(x_{n+1},X^*)^2\leq\left\| x_{n+1} - x^*_p \right\|^2_2
    = &~ \left\| x_n - x^*_p \right\|^2_2 - 2 \alpha \langle x_n - x^*_p,  \nabla f(x_n) \rangle + \alpha^2 \left\| \nabla f(x_n) \right\|^2_2 \nonumber \\
    \leq &~ \left\| x_n - x^*_p \right\|^2 - 2 \alpha \mu \left\| x_n - x^*_p \right\|^2_2 + \alpha^2 L^2 \left\| x_n - x^*_p \right\|^2_2 \nonumber \\
    = &~ \left( 1 - 2\mu\alpha + L^2\alpha^2 \right) \left\| x_n - x^*_p \right\|^2_2 \\
    = &~ \left( 1 - \frac{1}{\kappa^2} \right) d(x_n, X^*)^2 \tag*{for $\alpha = \frac{\mu}{L^2}$.}
\end{align}

\paragraph{\underline{Under $\SC^-(\mu)$ and $\QG^+(L)$}}

Assume $f \in \SC^-(\mu) \cap \QG^+(L)$ with some $L \geq \mu > 0$.
Note that this implies that $f$ has a unique minimum $x^*$.

Define $g(x) = \frac{1}{2} \| x - x^* \|_2^2 - \frac{1}{L} (f(x) - f^*)$; then, $g\in C^1(\R^d)$ and $g(x) \geq 0 = g^*$, since $f \in \QG^+(L)$, with $g(x^*) = 0$. 
Let $X^*$ be the set of all minima of $g$, including the $f$ minimizer $x^*$.

$g \in \SC^+\left(1 - \frac{1}{\kappa}\right)$ with $\kappa = \frac{\mu}{L}$: indeed, $\forall \, x,y \in\R^d$
\begin{gather}
g(y) - g(x) - \langle \nabla g(x), y-x\rangle  \nonumber \\
= \frac{1}{2} \| y - x^* \|_2^2 - \frac{1}{L} (f(y) - f^*) - \frac{1}{2} \| x - x^* \|_2^2 +  \frac{1}{L} (f(x) - f^*) - \langle (x-x^*) -\dfrac{1}{L} \nabla f(x) , y-x\rangle \nonumber \\
\leq -\dfrac{\mu}{2L} \|x-y\|_2^2 + \dfrac{1}{2} \le( \| y - x^* \|_2^2 -  \| x - x^* \|_2^2 - 2 \langle x-x^*,  y-x\rangle  \ri) \nonumber \\
\leq  -\dfrac{\mu}{2L} \|x-y\|_2^2 + \dfrac{1}{2} \le( \| y - x^* \|_2^2 +  \| x - x^* \|_2^2 - 2 \langle x-x^*, y-x\rangle  \ri) \nonumber \\
= \dfrac{1}{2}\le( 1 -\dfrac{\mu}{L}\ri) \|x-y\|_2^2 
\end{gather}
since $f \in \SC^-(\mu)$. This implies $g \in \EB^+\left(1 - \frac{1}{\kappa}\right)$:
$$ \| \nabla g(x) \|_2 = \left\| (x - x^*) - \frac{1}{L}\nabla f(x) \right\|_2 \leq \left(1 - \frac{1}{\kappa}\right) d(x, X^*) \leq \left(1 - \frac{1}{\kappa}\right) \|x - x^*\|_2$$
Therefore, in the GD algorithm with step size $\alpha = \frac{1}{L}$, we get
\begin{gather}
\| x_{n+1} - x^* \|_2 = \left\| x_n - x^* - \frac{1}{L}\nabla f(x_n) \right\|_2 \leq \left(1 - \frac{1}{\kappa}\right) \| x_n - x^* \|_2
\end{gather} 
Hence the linear rate $\left(1 - \frac{1}{\kappa}\right)^2$.

\paragraph{Rates of convergence for any pair of upper and lower condition.} 

We collected all the above results in Table \ref{convergence_table_2}. For any pair of upper and lower condition $f \in \mathcal{C}^+(L)\cap \mathcal{C}^-(\mu)$, we define $\kappa = \frac{L}{\mu}$. We will justify here all the entries.

The rates in the first column ($f \in \SC^-(\mu)$) follows from the fact that if $f \in \SC^+(L)$, we recover the classical convergence rate for $L$-smooth and $\mu$-strongly convex functions, while for any other upper condition $\mathcal{C}^+(L)$, we use the fact that $\mathcal{C}^+(L) \subseteq \QG^+(L)$ and we have convergence rate of $(1-\frac{1}{\kappa})^2$.

In the first row ($f \in \SC^+(L)$), the rate of convergence $1-\frac{1}{\kappa}$ holds for $f \in \PL^-(\mu)$ (as proven) and $f \in \, ^*\!\SC^-(\mu)$ (since $^*\!\SC^-(\mu) \subset  \PL^-(\mu)$); the rate of convergence $1-\frac{1}{\kappa^2}$ instead holds for  $f \in \EB^-(\mu)$ (since $\EB^-(\mu)\cap \SC^+(L) \subset \PL^-(\frac{\mu^2}{L})$) and consequently also for $f \in ~\RSI^-(\mu)$ (since $\RSI^-(\mu) \subset \EB^-(\mu)$).

We proved that for $f \in \RSI^-(\mu) \cap \EB^+(L)$ the GD algorithm converges with rate $1-\frac{1}{\kappa^2}$; the same rate of convergence is also valid for $f \in ~^*\!\SC^-(\mu) \subset \RSI^-(\mu)$ and/or $f \in \PL^+(L) \subset \EB^+(L)$. This justifies entries $(2,4)$, $(3,2)$ and $(3,4)$ in Table \ref{convergence_table_2}.

For entry $(2,2)$, we proved a convergence rate of $1-\frac{1}{\kappa}$ under assumption $f \in ~^*\!\SC^-(\mu) \cap \PL^+(L)$.
We also completed the entry $(2,4)$ under star convexity.
Since $\SC^+(L) \subset \PL^+(L)$, this rate also holds in $(1,4)$.

Similarly, under the additional assumption of star convexity, we have that $f \in \QG^-(\mu)\cap ~^*\!\SC(0) \subset \RSI^-(\frac{\mu}{2})$, therefore if $f \in  \QG^-(\mu)\cap ~^*\!\SC(0)  \cap \EB^+(L)$, GD converges with linear rate $1-\frac{1}{4\kappa^2}$. Following the same argument, for $f \in  \QG^-(\mu)\cap ~^*\!\SC(0)$ and upper conditions $f\in \PL^+(L)$ or $f\in \SC^+(L)$, GD converges with linear rate $1-\frac{1}{4\kappa}$. Entries $(2,3)$, $(2,5)$, $(3,3)$ and $(3,5)$ follows from $\PL^-(\mu) \subset \QG^-(\mu)$ and $\EB^-(\mu) \cap \QG^+(L) \subset \PL^-(\frac{\mu^2}{L})$.

If we assume $f$ to be convex, the rates of convergence on the fourth line ($f \in ~^*\!\SC^+(L)$) follow from the fact that $^*\!\SC^+(L) \cap \SC^-(0) \subset \EB^+(L)$. The rates on the last line ($f \in \QG^+(L)$) follow from $\QG^+(L) \cap \SC^-(0) \subset \EB^+(2L)$; similarly on the fifth line ($\RSI^+(L) \subset \QG^+(L)$).

\begin{table}[h!]
    
    \label{convergence_table_2}
\end{table}

As a last remark, we show that the additional assumption of $f$ being convex (or star convex) is fundamental in some cases in order to obtain convergence of the GD algorithm. 
We will show here that the sole pair of conditions $\SC^+(L) \cap \QG^-(\mu)$ doesn't guarantee convergence of gradient descent.

Let $\varepsilon,\eta>0$.
Consider the following function $f\in C^1(\R)$:
\begin{gather}
f(x) = \begin{cases}
        \frac{1}{2}x^2 &  x<1 \\
        -\frac{1}{2\varepsilon}x^2 + \frac{1+\varepsilon}{\varepsilon}x - \frac{1+\varepsilon}{2\varepsilon} & 1 \leq x < 1 + \varepsilon \\
        \frac{1 + \varepsilon}{2} &  1 + \varepsilon \leq x < 1 + \varepsilon + \eta \\
        \frac{1}{2}x^2 - \left(1 + \varepsilon + \eta \right) x + \frac{\left(1 + \varepsilon + \eta \right)^2}{2} + \frac{1+\varepsilon}{2} & 1 + \varepsilon + \eta \leq x
    \end{cases}
\end{gather}

By inspecting its second derivative (where defined) we can conclude that $f \in \SC^+(1)\cap \SC^-\left( -\frac{1}{\varepsilon} \right)$.

Furthermore, $\frac{2f(x)}{x^2}$ reaches its minimum at $\bar x = \frac{\left( 1 + \varepsilon + \eta \right)^2 + 1 + \varepsilon}{1 + \varepsilon + \eta}$,
with  $\frac{2f(\bar x)}{\bar x^2} = \frac{1 + \varepsilon}{\left( 1 + \varepsilon + \eta \right)^2 + 1 + \varepsilon} > 0$, therefore $f \in \QG^-\left( \frac{1 + \varepsilon}{\left( 1 + \varepsilon + \eta \right)^2 + 1 + \varepsilon} \right)$.

On the other hand, $f'(x) = 0 $ on $[1+\varepsilon, 1+\varepsilon+\eta]$, therefore if one of the iterates $x_j$ of the GD algorithm falls into this interval, then $x_{k} \in [1+\varepsilon, 1+\varepsilon+\eta]$ $\forall \, k\geq j$ and the algorithm fails to converge.

In the following, we will see sublinear convergence analysis under only upper conditions.

\paragraph{\underline{Under $~^*\!\SC^-(0)$ and $\SC^+(L)$}}

This proof is a very classical one \citep{bansal2017potential}, and it is based on studying the monotonic properties of the Lyapunov function $V_n = n\left(f(x_n) - f^*\right) + \frac{1}{2\alpha}d(x_n, X^*)^2$.
$\forall \, n\in \mathbb{N}$, let $x_{n,p}^*$ be the projection of $x_n$ onto $X^*$.

\begin{align}
    V_{n+1} = &~ (n+1)\left(f(x_{n+1}) - f^*\right) + \frac{1}{2\alpha}\|x_{n+1} - x_{n,p}^*\|^2 \nonumber \\ \overset{\SC^+(L)}{\leq} &~ (n+1)\left(f(x_n) - f^* + \left( \frac{L}{2}\alpha^2 - \alpha \right) \left\| \nabla f(x_n) \right\|^2 \right) \nonumber \\
    &~ + \frac{1}{2\alpha}\left( \|x_n - x_{n,p}^*\|^2 - 2\alpha \langle \nabla f(x_n), x_n - x_{n,p}^* \rangle + \alpha^2 \left\| \nabla f(x_n) \right\|^2 \right) \nonumber \\
    = &~ V_n + \left( f(x_n) - f^* \right) + \left( (n+1)\left( \frac{L}{2}\alpha^2 - \alpha \right) + \frac{\alpha}{2} \right) \left\| \nabla f(x_n) \right\|^2 - \langle \nabla f(x_n), x_n - x_{n,p}^* \rangle \nonumber \\
    \overset{~^*\!\SC^-(0)}{\leq} &~ V_n + \left( (n+1)\left( \frac{L}{2}\alpha^2 - \alpha \right) + \frac{\alpha}{2} \right) \left\| \nabla f(x_n) \right\|^2 \nonumber \\
    \leq &~ V_n \tag*{for $\alpha = \frac{1}{L}$}
\end{align}

Therefore, $V_n$ is decreasing and in particular
\begin{equation}
    n(f(x_n) - f^*) \leq V_n \leq V_0 \leq \frac{L}{2}d(x_0, X^*)^2
\end{equation}

Leading to the desired rate

\begin{equation}
    f(x_n) - f^* \leq \frac{L}{2n}d(x_0, X^*)^2
\end{equation}

\paragraph{\underline{Under $~^*\!\SC^-(0)$ and $\PL^+(L)$}}

$\forall \, n\in \mathbb{N}$, let $x_{n,p}^*$ be the projection of $x_n$ onto $X^*$.

\begin{align}
    d(x_{n+1}, X^*)^2 &\leq \|x_{n+1} - x_{n,p}^*\|^2 =  \|x_n - x_{n,p}^*\|^2 - 2\alpha \langle \nabla f(x_n), x_n - x_{n,p}^* \rangle + \alpha^2 \left\| \nabla f(x_n) \right\|^2 \nonumber \\
    &\leq  \|x_n - x_{n,p}^*\|^2 - 2\alpha (f(x_n) - f^*) + \alpha^2 \times 2L(f(x_n) - f^*)
\end{align}
therefore,
\begin{gather}
        2\alpha(1 - L\alpha)(f(x_n) - f^*) \leq  d(x_n, X^*)^2 - d(x_{n+1}, X^*)^2.
\end{gather}

By summing the inequality above for  $k=0,\ldots, n$, we have
\begin{align}
    2\alpha(1 - L\alpha)\sum_{k=0}^n(f(x_k) - f^*) \leq &~ d(x_0, X^*)^2 - d(x_{n+1}, X^*)^2 \leq d(x_0, X^*)^2
\end{align}
and taking $\alpha = \frac{1}{2L}$,
\begin{align}
    \frac{1}{n+1} \sum_{k=0}^n(f(x_k) - f^*) \leq &~ \frac{2L}{n+1}d(x_0, X^*)^2
\end{align}
we can conclude
\begin{align}
    \min_{k\in[|0, n|]}(f(x_k) - f^*) \leq &~ \frac{2L}{n+1}d(x_0, X^*)^2
\end{align}

If additionally $f\in \SC^-(0)$ (convex), we have the stronger result
\begin{align}
    f\left(\frac{1}{n+1}\sum_{k=0}^n x_k \right) - f^* \leq &~ \frac{2L}{n+1}d(x_0, X^*)^2.
\end{align}

    \section{Adaptive step size and application to logistic regression}\label{App_log_regr}
    Let $f \in C^1(\R^d)$ be a function to optimize, and let $g \in C^1\left( [ f^*, +\infty) \right)$ be an increasing function.
It is easy to see that finding the minimum of $g \circ f$ is equivalent to finding the minimum of $f$, and a GD algorithm with constant step size on $g \circ f$ leads to a GD algorithm on $f$ with adaptive step size:
\begin{equation}
    \label{eq:constant_to_adaptive_step_size}
    x_{n+1} = x_n - \alpha \nabla \left( g \circ f \right)(x_n) \quad \Leftrightarrow \quad x_{n+1} = x_n - \alpha g'(f(x_n)) \nabla f(x_n).
\end{equation}

We briefly recall here the definition of the $\Theta$ notation, because it will be occasionally used in the following proposition and proof in order to preserve their readability.
\begin{Def}[{\bf $\Theta$ notation}]
Given two functions $f,g \in C^0(\R)$, $g \geq 0$, we say that 
$$f(x) \in \Theta\le(g(x) \ri) \qquad \text{as } x \to x_0$$
if $\exists \, \delta, m, M >0$ such that $\forall \, x$ with $0<|x-x_0|<\delta$:
\begin{gather}
m \, g(x) \leq \le| f(x) \ri| \leq M \, g(x). 
\end{gather}
Similarly, we say that 
$$f(x) \in \Theta\le(g(x) \ri) \qquad \text{as } x \to +\infty$$
if $\exists \, K, m, M >0$ such that $\forall \, x >K$:
\begin{gather}
m \, g(x) \leq \le| f(x) \ri| \leq M \, g(x). 
\end{gather}
\end{Def}

\begin{Prop}
\label{prop:use_of_QG_for_adaptive}
    Given $f \in C^1\left( \mathbb{R}^d \right)$, assume that 
    \begin{align}
    &f(x) - f^* \in \Theta\le( d(x, X^*)^{\beta} \ri)&  \text{as } d(x,X^*) \rightarrow 0,\\
    &f(x) - f^* \in \Theta \le( d(x,X^*)^{\gamma} \ri) & \text{as } d(x,X^*) \rightarrow \infty,
    \end{align} 
    for some $\beta, \gamma \in (0, \infty)$.
    Consider the functions
    \begin{align}
    \begin{array}{cc}
    g: (-c,+\infty) \to \R_+ \qquad&\qquad h: [f^*,+\infty) \to \R_+\cup \{0\}\\
    u \mapsto  \le( u + c \ri)^{\frac{\beta}{\gamma}} \qquad&\qquad t \mapsto (t-f^*)^{\frac{2}{\beta}} 
    \end{array}
    \end{align} 
    where $c>0$ is an arbitrary positive constant. 
    Then, $g \circ h \circ f \in \QG^-(\mu) \cap \QG^+(L)$ for some $\mu, L >0$.
\end{Prop}

In the case $g \circ f$ is convex, we obtain a linear rate convergence from Table 1. 
This is easily satisfied when $f$ is convex and $\beta, \gamma \in (0, 2]$.

\begin{Rem}
    This property leads to an adaptive step size $\tilde \alpha_n = \alpha g'(f(x_n))$ for the adaptive GD algorithm which requires the knowledge of the precise value of $f^*$.
    However, in the particular case where $\beta = 2$ and $f^* > 0$, we can take $c = f^*$ and obtain a step size $\tilde \alpha_n = \alpha \frac{2}{\gamma} f(x_n)^{\frac{2}{\gamma}-1} $. 
\end{Rem}

\begin{proof} $g\in C^1((-c,+\infty)))$ and $g(u) > 0$ on its domain. It is easy to see that
\begin{align}
&g(u) - c^{\frac{\beta}{\gamma}} \in \Theta\le( u\ri) & \text{as } u\to 0\\
&g(u) - c^{\frac{\beta}{\gamma}} \in \Theta \le(u^{\frac{\beta}{\gamma}} \ri) & \text{as  }u \to + \infty
\end{align}

Consider the function $h(f(x)) = (f(x)-f^*)^{\frac{2}{\beta}}$: clearly, $h\circ f$ is continuous on $\R^d$ ($f$ is continuous) and $h(f(x)) =0 \Leftrightarrow x \in X^*$.
    By continuity of all the functions involved, $\exists \, \delta, m_0, M_0 >0$ such that 
    \begin{gather}
        m_0\leq \dfrac{g( h(f(x))) - g ( h(f^*))}{h(f(x))} = \frac{\le( (f(x) - f^*)^{\frac{2}{\beta}} + c \ri)^{\frac{\beta}{\gamma}} -c^{\frac{\beta}{\gamma}}}{ (f(x) - f^*)^{\frac{2}{\beta}}} \leq M_0 \qquad  \text{for } 0<d(x,X^*)< \delta 
    \end{gather}
and using the fact that $f(x) - f^* \in \Theta\le( d(x, X^*)^{\beta} \ri)$ as $d(x, X^*)\to 0$
    \begin{gather}
        \tilde m_0\leq  \dfrac{\le( (f(x) - f^*)^{\frac{2}{\beta}} + c \ri)^{\frac{\beta}{\gamma}} -c^{\frac{\beta}{\gamma}}}{ d(x, X^*)^2 } \leq \tilde M_0 \qquad  \text{for } 0<d(x,X^*)< \delta 
    \end{gather}
    i.e. $g(h(f(x))) - g(h(f^*)) \in \Theta\le( d(x,X^*)^2\ri)$.

    Similarly, $\exists \, K, m_\infty, M_\infty >0$ such that
    \begin{gather}
m_\infty \leq \dfrac{g (h(f(x))) - g(h(f^*))}{h(f(x))^{\frac{\beta}{\gamma}}} =
  \frac{\le( (f(x) - f^*)^{\frac{2}{\beta}} + c \ri)^{\frac{\beta}{\gamma}} -c^{\frac{\beta}{\gamma}}}{ (f(x) - f^*)^{\frac{2}{\gamma}} } \leq M_\infty \qquad \text{for } d(x,X^*)>K
    \end{gather}
and using the fact that $f(x) - f^* \in \Theta\le( d(x, X^*)^{\gamma} \ri)$ as $d(x, X^*)\to\infty $
    \begin{gather}
        \tilde m_\infty\leq \frac{\le( (f(x) - f^*)^{\frac{2}{\beta}} + c \ri)^{\frac{\beta}{\gamma}} -c^{\frac{\beta}{\gamma}}}{ d(x, X^*)^{2} } \leq \tilde M_\infty \qquad  \text{for } d(x,X^*)> K
    \end{gather}
    i.e. $g(h(f(x))) - g(h(f^*)) \in \Theta\le( d(x,X^*)^2\ri)$.

In conclusion,  $\exists \, R >0$, $\exists \, \mu_1, \mu_2, L_1, L_2 >0$ such that
    \begin{align}
        & \mu_1 \leq \frac{g ( h(f(x))) - g ( h(f^*))}{d(x,X^*)^{2}} \leq L_1  \qquad  \text{for } 0<d(x,X^*) \leq R\\
        & \mu_2 \leq \frac{g ( h(f(x))) - g ( h(f^*))}{d(x,X^*)^{2}} \leq L_2  \qquad \text{for } d(x,X^*) > R
    \end{align}
    By setting $\mu = \min \{\mu_1,\mu_2\}$ and $L = \max\{L_1,L_2\}$, we have $g \circ h\circ f \in \QG^-(\mu) \cap \QG^+(L) $.
\end{proof}

\paragraph{Logistic regression: settings and notations}

Logistic regression is a common ML tool that is well studied and documented (see e.g. \citep{Bach2013} and \citep{BachMoulines}).

Given a distribution of data $X \sim \mathcal{D}$, and their class $Y \in \lbrace -1, 1 \rbrace$, logistic regression aims at finding the maximum likelihood of the parametrized set of distributions verifying that $\ln{\frac{\mathbb{P}\left[ Y = 1 | X \right]}{1 - \mathbb{P}\left[ Y = 1 | X \right]}}$ is linear in $X$.
We call $\omega$ the associated coefficient.

\begin{equation}
    \ln{\frac{\mathbb{P}\left[ Y = 1 | X \right]}{1 - \mathbb{P}\left[ Y = 1 | X \right]}} =  \langle \omega , X \rangle  \label{eq:logistic_basis}
\end{equation}

Note the bias can be included in $\omega$ by adding an additional dimension to $X$ whose coordinate would always be 1.
Eq.\eqref{eq:logistic_basis} is equivalent to

\begin{equation}
    \mathbb{P}\left[ Y = 1 | X \right] = \sigma \left( \langle  \omega , X\rangle  \right)
\end{equation}

with $\sigma (x) =  \frac{1}{1 + e^{-x}}$.

Then the likelihood of $Y | X$ is $\mathbb{P}\left[ Y = 1 | X \right]^{\mathbf{1}_{Y = 1}}\mathbb{P}\left[ Y = -1 | X \right]^{\mathbf{1}_{Y = -1}}$.
We aim at maximizing the log-likelihood (equivalently minimizing its opposite)
\begin{align}
    f(\omega)
    = &~
    - \mathbb{E} \left[
    \mathbf{1}_{Y = 1}
    \ln \sigma \left(  \langle \omega , X\rangle   \right)
    +
    \mathbf{1}_{Y = -1}
    \ln \sigma \left( - \langle \omega , X\rangle  \right)
    \right] \nonumber \\
    = &~
    - \mathbb{E} \left[
    \ln \sigma \left( Y \,\langle \omega , X\rangle  \right)
    \right] \nonumber \\
    = &~
    - \mathbb{E}_{Z \sim YX} \left[
    \ln \sigma \left( \langle \omega , Z \rangle \right)
    \right].
\end{align}

The function $f(\omega)$ satisfies:
\begin{align}
    &f(\omega)
    = 
    \mathbb{E} \left[
    - \ln \sigma \left( \langle \omega , Z\rangle  \right) \right] \\
    &\nabla f(\omega)
    =
    \mathbb{E} \left[
    - (1 - \sigma) \left(\langle  \omega , Z\rangle  \right) Z \right] \\
    &\nabla^2 f(\omega)
    =
    \mathbb{E} \left[
    \sigma (1 - \sigma) \left(\langle  \omega , Z\rangle  \right) Z Z^\top  \right]
\end{align}

\begin{Prop}
\label{prop:all_about_logistic_regression}
    Under the following assumptions:
    \begin{align}
        & \mathbb{P}\left[  \langle \omega , Z \rangle  > 0 \right] > 0, \qquad \forall\,  \omega \neq 0 \label{hyp:logisticspan} \\
        & \mathbb{E}\left[ \|Z\|^2_2 \right] < \infty \label{hyp:logisticL2}
    \end{align}
    the logistic regression function $f$ is positive, smooth and (strictly) convex on $\R^d$; automatically, as described in \citep{DBLP:journals/corr/KarimiNS16}, it is strongly convex on any compact $\mathcal{K} \subset \R^d$. Additionally, $f$ grows linearly at infinity.
    
\end{Prop}

Note that the the hypothesis \eqref{hyp:logisticL2} is verified for discrete measure as in practice.
The hypothesis \eqref{hyp:logisticspan} ensures there is enough disparity in the data.

\begin{proof}
By construction, $f(\omega)$ is the expectation of a positive variable, therefore $f(\omega) > 0$ $\forall \, \omega \in \R^d$.

Let $r \in \R^d$ be a unit vector ($\|r\|_2=1$), then $\forall \, \omega \in \R^d$
\begin{align}
            r^\top \nabla^2 f(\omega) r
            &= 
            \mathbb{E} \left[
            \sigma (1 - \sigma) \left( \langle \omega , Z\rangle  \right)  \langle Z , r\rangle ^2 \right] 
            \leq
            \mathbb{E} \left[
            \sigma (1 - \sigma) \left( \langle  \omega , Z \rangle \right) \left\| Z \right\|_2^2 \right] \nonumber \\
            &\leq 
            \mathbb{E} \left[ \left\| Z \right\|_2^2 \right] < \infty 
        \end{align}
        thanks to  \eqref{hyp:logisticL2}. Therefore, $\exists \, M >0$ such that 
        $ M\, \mathrm{I}_d - \nabla^2 f(\omega) $ is positive semi-definite, i.e. $f$ is smooth.
        
        Additionally,  $\forall \, r \in\R^d$ unit vector, $\forall \, \omega \in \R^d$
\begin{gather}
    r^\top \nabla^2 f(\omega) r
            = 
            \mathbb{E} \left[
            \sigma (1 - \sigma) \left(  \langle \omega , Z \rangle \right)  \langle Z , r \rangle^2 \right]
            >  0
\end{gather}
thanks to \eqref{hyp:logisticspan}, i.e. $f$ is strictly convex. Furthermore, for any $\mathcal{K} \subset \R^d$ compact, $f$ is strongly convex on $\mathcal{K}$.

On the other hand, it is not strongly convex on the full space $\R^d$ and $f \notin \QG^-(\mu)$ for any $\mu\geq 0$, as it grows linearly in infinity: $f(\omega) \in \Theta(\left\| \omega \right\|_2)$, as $\|\omega\|_2 \to +\infty$.
        
        Indeed, $\forall \, t \in \R$
        \begin{align*}
             \ln{\sigma(t)}= \ln\left(1 + e^{-t}\right)
            \in \left[\max\{0, -t\}, \ln(2) + \max\{0, -t\} \right],
        \end{align*}
        therefore, $\forall \, \omega \in \R^d$, $\mathbb{E}\left[ \max\{0, -\langle \omega , Z \rangle \} \right]
        \leq f(\omega) \leq
        \ln(2) + \mathbb{E}\left[ \max\{0, -\langle  \omega , Z \rangle \} \right]$.
        
        On the one hand, 
        \begin{align}
            f(\omega) \leq &~ \ln(2) + \mathbb{E}\left[ \max\{0, -\langle \omega , Z\rangle \} \right] 
            \leq  \ln(2) + \mathbb{E}\left[ \|\omega\|_2 \| Z \|_2 \right] \nonumber \\
            \leq & \ln(2) + \|\omega\|_2 \sqrt{\mathbb{E}\left[ \| Z \|^2_2 \right]} \leq \ln(2) + K_1 \|\omega\|_2
        \end{align}
        for some $K_1 >0$, thanks to \eqref{hyp:logisticL2}.
        On the other hand,
        \begin{gather}
            f(\omega) \geq \mathbb{E}\left[ \max\{0, - \langle \omega , Z\rangle \} \right] 
            \geq K_2  \|\omega\|_2
        \end{gather}
        
        where $K_2 = \underset{\|\omega\|_2 = 1}{\min} \mathbb{E}\left[ \max\{0, - \langle \omega , Z\rangle \} \right]$. 
        
        It remains to prove that $K_2>0$. Note that the sphere $S^{d-1} \in \mathbb{R}^d$ is a compact set. Hence any continuous function defined on the sphere reaches its minimum and it is clear that $\omega \mapsto \mathbb{E}\left[ \max\{0, - \langle \omega , Z\rangle \}\right]$ is Lipschitz continuous hence continuous. Then we only need to show that for any $\omega$ with norm 1, we have $\mathbb{E}\left[ \max\{0, -\langle \omega , Z\rangle \} \right]>0$.
        
        We prove the latest by contradiction.
        Assume $\|\omega\|_2 = 1$ and $\mathbb{E}\left[ \max\{0, -\langle \omega , Z\rangle \} \right]=0$.
        Since the integrand is non negative, and the integral is 0, the integrand has to be 0 almost surely (i.e. with probability 1). We have $\mathbb{P}\left[ - \langle \omega , Z \rangle\leq 0 \right]=1$, or again $\mathbb{P}\left[ -\langle \omega , Z\rangle > 0 \right]=0$, which contradicts \eqref{hyp:logisticspan}.
\end{proof}

We conclude that the logistic regression is strongly convex and smooth on every compact set; therefore for any compact set $\mathcal{K} \subset \R^d$ and for any $x_0\in \mathcal{K}$, one can fine-tune the GD algorithm starting in $x_0$ such that it converges linearly. 
However, the logistic regression is not strongly convex on the full space $\R^d$, and global uniform tuning of GD for linear convergence rate is not provided by classical studies of GD algorithm on strongly convex and smooth functions.

On the other hand, $f(\omega)$  verifies all the assumptions of Proposition \ref{prop:use_of_QG_for_adaptive} with $\beta = 2$ and $\gamma = 1$ and $f$ is convex. 
Therefore, we can have linear rate of convergence of GD algorithm on the function $g \circ f$ where  $g(t) = (t - f^* + c)^2$, for any $c>0$. In particular, since $f$ is positive, we choose $c=f^*$: then, thanks to Proposition \ref{prop:use_of_QG_for_adaptive}, we have linear convergence rate of GD on the function $f^2(\omega) \in \QG^-(\mu) \cap \QG^-(L)$ for some $\mu,L>0$ (see Table \ref{convergence_table}), and the exact knowledge of $f^*$ is not required.



In summary, classical studies of GD with constant step size  don't allow to find an optimal global (i.e. independent on the initialization $x_0$) step size $\alpha$ so that GD algorithm (linearly) converges on $f$. However, from the study above, we showed that a linear rate convergence can be achieved with an adaptive step size $\tilde  \alpha_n = \alpha  f(x_n)$ for well tuned $\alpha$ (according to the upper and lower properties of $f$), regardless of the initialization $x_0$.

\end{document}